%% file: sample-sigconf.tex
\documentclass[sigconf]{acmart}
\AtBeginDocument{%
  }

\usepackage{balance}
\usepackage{algorithm}
\usepackage{algorithmic}
\usepackage{textcomp}
\usepackage[utf8]{inputenc} 
\usepackage[T1]{fontenc}    
\usepackage{booktabs}       
\usepackage{amsfonts}       
\usepackage{hyphenat}       
\usepackage{amsmath}        
\usepackage{nicefrac}       
\usepackage{microtype}      
\usepackage{xcolor}         
\usepackage{colortbl}       
\usepackage{graphicx}
\usepackage{makecell}
\usepackage{multirow}
\usepackage{multicol}
\usepackage{booktabs,multirow}
\usepackage{array} 
\usepackage{scalefnt}
\usepackage{subcaption}
\usepackage{float}
\usepackage{cleveref}

\usepackage{placeins}
\usepackage{newfloat}
\usepackage{listings}
\usepackage{xcolor}
\definecolor{myblue}{HTML}{1D7AB8}
\definecolor{myred}{HTML}{E63946}

\copyrightyear{2026}
\acmYear{2026}
\setcopyright{cc}
\setcctype{by}
\acmConference[CIKM '26]{Proceedings of the 35th ACM International Conference on Information and Knowledge Management}{November 07--11, 2026}{Rome, Italy}
\acmBooktitle{Proceedings of the 35th ACM International Conference on Information and Knowledge Management (CIKM '26), November 07--11, 2026, Rome, Italy}
\acmDOI{10.1145/3799682.3840947}
\acmISBN{979-8-4007-2539-5/2026/11}

\begin{document}

\title{UPLOTS: A Unified Pretrained Language Model for Constrained Time-series Generation}


\author{Du Yin}
\orcid{0000-0002-2345-0683}
\email{du.yin@unsw.edu.au}
\affiliation{%
  \institution{The University of New South Wales}
  \city{Sydney}
  \country{Australia}
}

\author{Hao Xue}
\orcid{0000-0003-1700-9215}
\email{haoxue@hkust-gz.edu.cn}
\affiliation{%
  \institution{The Hong Kong University of Science and Technology Guangzhou (HKUST-GZ)}
  \city{Guangzhou}
  \country{China}
}

\author{Jinliang Deng}
\orcid{0000-0002-0759-947X}
\email{dengjinliang@ust.hk}
\affiliation{%
  \institution{The Hong Kong University of Science and Technology}
  \country{Hong Kong}
}

\author{Yang Yang}
\orcid{0000-0003-3960-5739}
\email{yang.yang26@unsw.edu.au}
\affiliation{%
  \institution{The University of New South Wales}
  \city{Sydney}
  \country{Australia}
}

\author{Shuang Ao}
\orcid{0009-0008-3321-8416}
\email{shuang.ao@unsw.edu.au}
\affiliation{%
  \institution{The University of New South Wales}
  \city{Sydney}
  \country{Australia}
}

\author{Arian Prabowo}
\orcid{0000-0002-0459-354X}
\email{arian.prabowo@gmail.com}
\affiliation{%
  \institution{The University of New South Wales}
  \city{Sydney}
  \country{Australia}
}

\author{Flora Salim}
\orcid{0000-0002-1237-1664}
\email{flora.salim@unsw.edu.au}
\affiliation{%
  \institution{The University of New South Wales}
  \city{Sydney}
  \country{Australia}
}

\renewcommand{\shortauthors}{Du Yin et al.}

\begin{abstract}
Many time-series generation studies train a separate generator for each dataset. We study a complementary setting in which one shared, channel-independent generator is trained on several timestamp-derived scenario pools and a compact textual prompt selects the desired pool at inference. We propose UPLOTS, a \textbf{U}nified, \textbf{P}rompt-guided \textbf{L}anguage model framework f\textbf{O}r constrained \textbf{T}ime-\textbf{S}eries generation. UPLOTS combines a Diffusion-TS generator with a truncated pretrained language-model backbone, a Time-series Prompt Embedding Module, and two multi-dataset balancing mechanisms: Curriculum-Aware Loss Modulation and Rolling-Loss Dynamic Sampling. We evaluate four real-world datasets under morning- and evening-peak scenarios and extend the analysis to 14 prompt configurations on ETTh and Energy, covering temporal, calendar, load-level, and volatility subsets. The results support unified, in-distribution generation for the predefined datasets and scenario prompts. Downstream forecasting experiments further assess the utility of the generated samples for data augmentation. Our code is available at \url{https://github.com/cruiseresearchgroup/UPLOTS}.

\end{abstract}




\maketitle

\section{Introduction}
Time-series generation plays a crucial role in diverse applications spanning finance, transportation modeling~\cite{yin2025xxltraffic}, and healthcare analytics. Existing methods~\citep{ang2023tsgbench} typically rely on designing or training a distinct model for each specific dataset or task, resulting in limited scalability and inefficient exploitation of common temporal structures across various domains. This fragmentation limits the model's capability to assimilate extensive knowledge from varied datasets, thereby restricting its performance in time-series generation tasks.

\begin{figure*}[!ht]
\centering
\includegraphics[width=0.99\linewidth]{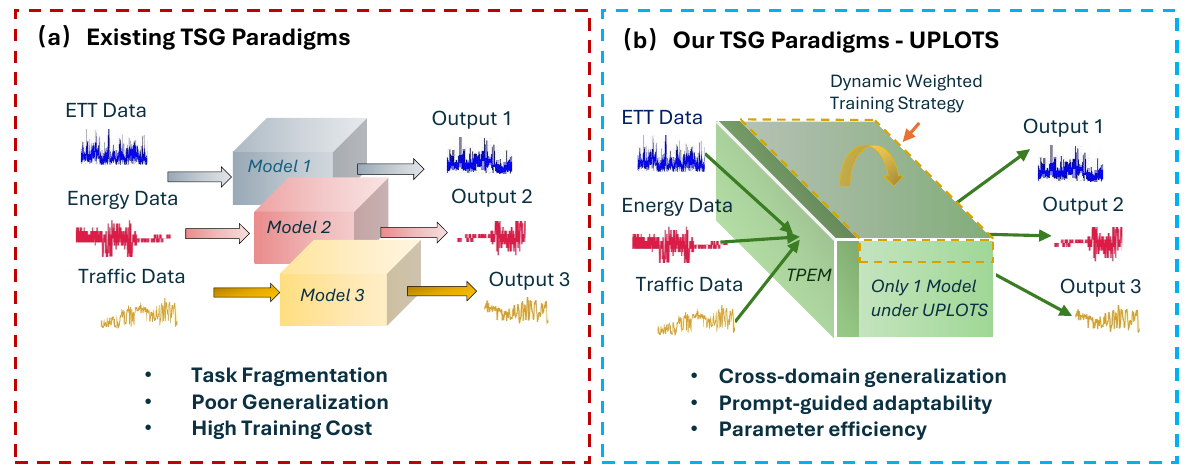}
\caption{
\textbf{Conceptual comparison between dataset-specific TSG pipelines and UPLOTS.} The left (red frame) depicts separately trained generators. The right (blue frame) depicts one shared training pipeline for multiple predefined dataset--scenario pools. The Time-series Prompt Embedding Module (TPEM, green block) conditions generation on the selected prompt.}
\Description{A side-by-side schematic compares three dataset-specific generators with UPLOTS, which shares one prompt-conditioned generator across ETT, Energy, and traffic data.}

\label{fig:intro}
\end{figure*}

Recent VAE, GAN, and diffusion-based models have catalyzed notable progress in time-series generation, expanding the repertoire of generative mechanisms and delivering richer, more flexible representations of complex temporal dynamics.
Despite continued advances in VAE-, GAN-, and diffusion-based generation, many evaluations retain the one-model-per-dataset setup illustrated on the left of Figure~\ref{fig:intro}. Training and maintaining separate generators becomes increasingly costly as the number of datasets and operational scenarios grows. TimeDP~\cite{huang2025timedp} is an important exception: it learns time-series prototypes and extracts a domain prompt from few-shot target-domain examples. Our setting is different and narrower. For a collection of predefined, seen dataset--scenario pools, the user supplies an explicit textual identifier and no target exemplar series is required at inference. We therefore focus on controllable scenario selection and balanced joint training, rather than claiming unseen-domain generation.

Inspired by prompt conditioning in language models, we combine a truncated pretrained transformer with Diffusion-TS. At each reverse step, the selected scenario identifier is fused with the current noisy sequence, and the resulting representation is passed to the Diffusion-TS encoder--decoder to estimate the clean sequence. The model is trained jointly over all selected pools and can switch among their prompts without retraining. The prompt changes the denoiser input at every step. It does not change the diffusion schedule or the sampler equations.

UPLOTS is illustrated on the right of Figure~\ref{fig:intro}. The GPT backbone is mostly frozen, while its LayerNorm and positional embeddings, the time-series and output projections, and the Diffusion-TS components remain trainable. We further introduce a Dynamic Weighted Training Strategy: the Curriculum-Aware Loss Modulator temporarily reduces the loss contribution of high-loss pools, while Rolling-Loss Dynamic Sampling allocates more micro-batches to pools with larger rolling loss.

\subsection{Contributions}

\begin{itemize}
  \item We formulate unified scenario-conditioned generation across several seen datasets and scenario pools. A shared channel-independent generator switches among these pools through compact textual identifiers without per-scenario retraining.
  \item We introduce TPEM for prompt--series fusion and a Dynamic Weighted Training Strategy that combines CALM loss weighting with RLDS micro-batch allocation. Both mechanisms are defined explicitly in terms of rolling per-pool losses.
  \item We evaluate generation fidelity and downstream utility on four real-world datasets and extend the analysis to 14 configurations on ETTh and Energy.

\end{itemize}

\section{Related Work}
\label{gen_inst}

\subsection{Time-series Generation}


\paragraph{VAE} C-RNN-VAE~\cite{fabius2014variational} fuses a VAE with conditional RNN decoders for coherent sequence synthesis. VRNN~\cite{chung2015recurrent} and SRNN~\cite{fraccaro2016sequential} enrich the latent space with recurrent and hierarchical dynamics. Later variants extend this line of work: GP-VAE~\cite{pmlr-v108-fortuin20a} adopts Gaussian-process priors, TimeVAE~\cite{desai2021timevae} adds a dual-branch decoder, and Time-Causal VAE~\cite{acciaio2024time} imposes causality with a RealNVP prior. The latest models—Koopman VAE~\cite{naiman2024generative}, TimeAutoDiff~\cite{suh2024timeautodiff}, TimeVQVAE~\cite{lee2023vector}, and Time-Based VQVAE~\cite{murad2025synthetic}—inject Koopman theory, diffusion, or vector-quantization mechanisms to further improve temporal fidelity. Although early VAEs often produced blurred sequences, recent advances such as KoVAE~\cite{naiman2024generative} and TimeVQVAE~\cite{lee2023vector} employ Koopman dynamics or vector quantization to deliver high-fidelity samples comparable to diffusion approaches.

\paragraph{GAN} C-RNN-GAN~\cite{mogren2016c} first validated adversarial training for continuous sequences, followed by LSTM-based RGAN/RCGAN~\cite{esteban2017real} for medical time-series and TimeGAN~\cite{yoon2019time}'s hybrid supervised-adversarial approach. Recent research works include GT-GAN~\cite{jeon2022gt} (global-local critics), PSA-GAN~\cite{jehapsa} (hierarchical self-attention), RTSGAN~\cite{pei2021towards} (encoder-decoder for irregular series), and COT-GAN~\cite{xu2020cot} (causal optimal transport).
GANs can produce realistic sequences but may suffer from unstable training, mode collapse, and poor long-range coherence.


\paragraph{DDPM} DDPM models generate sequences by iterative denoising: DiffWave~\cite{kongdiffwave} enforces monotonic or periodic structure, DiffTime~\cite{coletta2023constrained} enables conditional forecasting, Diffusion-TS~\cite{yuandiffusion} adds a transformer with disentangled temporal cues, WaveStitch~\cite{shankar2025wavestitch} accelerates sampling via segment stitching, D³M~\cite{yan2024probabilistic} fuses discrete diffusion with continuous flows, and DiffImp~\cite{gao2024diffimp} targets imputation with a bidirectional Mamba core. DDIM~\cite{song2021ddim} reduces the number of sampling steps through a non-Markovian implicit process while retaining the DDPM training objective. Sampler acceleration is orthogonal to the conditioning mechanism studied here.

Furthermore, SDformer~\cite{chen2024sdformer} discretizes series for Transformer decoding, and ImagenTime~\cite{naiman2024utilizing} converts sequences to images to reuse vision diffusers. FIDE~\cite{galib2024fide} combines frequency-domain inflation with block-maxima conditioning. TimeDP~\cite{huang2025timedp}, the closest multi-domain baseline, is label-free and text-free: it learns semantic prototypes and obtains a sampling-time domain prompt from few-shot target examples. UPLOTS instead uses predefined textual scenario identifiers for seen pools and does not require target examples at inference. T2S~\cite{ge2025t2s} studies a broader text-to-series setting with a length-adaptive VAE and Diffusion Transformer trained on a 600K+ corpus.

\subsection{LLM for Time-series Models}

LLMs now tackle time-series by “reading” numbers as text. Gruver et al.~\cite{gruver2023large} showed that digit-stream tokenization lets GPT-3~\cite{brown2020language} and LLaMA-2~\cite{touvron2023llama} deliver zero-shot forecasts competitive with bespoke models. Prompt engineering generalizes this idea: GPT4TS~\cite{zhou2023one} handles forecasting, classification, and anomaly detection. UniTime~\cite{liu2024unitime} unifies cross-domain tasks via template-guided fine-tuning. TEST~\cite{suntest} aligns numeric prototypes with LLM embeddings, and TimeLLM~\cite{jin2024time} reprograms frozen models through “Prompt-as-Prefix.” Scaling up, TimeGPT~\cite{garza2023timegpt} pretrains on approximately 100 billion time points, and CEHR-GPT~\cite{pang2024cehr} generates privacy-preserving EHR sequences. PaD-TS~\cite{Li_Meng_Bi_Urnes_Chen_2025} adds dual encoders and a population-preservation loss to diffusion models, while BRIDGE~\cite{li2025bridge} couples multi-agent LLM prompt refinement with diffusion-based generation. Together, these works span prompt-only, hybrid, and fully pretrained paradigms that extend LLMs to versatile, cross-domain time-series synthesis and analysis.

These studies show that language-model representations can be useful for time-series tasks, while their role in controlled generation remains less settled. UPLOTS tests whether a mostly frozen pretrained transformer can serve as a shared prompt-conditioning component inside a trainable diffusion generator. Section~\ref{abla} compares this default with same-width random initialization, end-to-end backbone training, learned scenario IDs, and removal of the prompt pathway.

\section{Preliminaries}

\subsection{Problem Definition}
\label{sec:problem}
Let \(\mathcal{D}_k=\{\mathbf{X}^{(i)}_k\}_{i=1}^{N_k}\), where
\(\mathbf{X}^{(i)}_k\in\mathbb{R}^{T\times C_k}\), denote the sequence segments assigned to a predefined dataset--scenario pool \(k\), and let \(p_k\) be its textual identifier. Because the datasets have different channel counts, we adopt a shared channel-independent formulation: each channel trajectory \(\mathbf{X}^{(i)}_{k,:,c}\in\mathbb{R}^{T\times 1}\) is treated as one training sequence, while model parameters are shared across all channels and pools. The objective is to learn
\begin{equation}\label{pre1}
  \hat{\mathbf{x}}\sim p_{\theta}(\mathbf{x}\mid p_k),
  \qquad \mathbf{x},\hat{\mathbf{x}}\in\mathbb{R}^{T\times 1},
\end{equation}
so that the generated channel-wise distribution matches the reference distribution associated with \(p_k\). At inference, selecting a seen prompt \(p_k\) changes the requested scenario without retraining the generator.

\subsection{Basic TS Generation Model: Diffusion-TS}
\label{diffts}
Diffusion-TS proceeds in three steps:
\paragraph{Forward diffusion} A clean sequence \(\mathbf{x}_{0}\!\in\!\mathbb{R}^{T\times 1}\) is progressively noised via
\(q(\mathbf{x}_{s}\mid\mathbf{x}_{0})=
\mathcal{N}(\sqrt{\bar\alpha_s}\,\mathbf{x}_{0},(1-\bar\alpha_s)\mathbf I)\),
for diffusion step \(s=1,\dots,S\) and cumulative variance schedule \(\{\bar\alpha_s\}\).
\paragraph{Reverse model and loss} The denoiser predicts the clean input, \(\hat{\mathbf{x}}_{0}=f_{\theta}(\mathbf{x}_{s},s,p_k)\). Training minimizes the same time/frequency reconstruction form used by Diffusion-TS:
\begin{equation}\label{diff2}
\begin{split}
\mathcal{L}_{\mathrm{diff}}={}&\rho(\mathbf{x}_{0},\hat{\mathbf{x}}_{0})\\
&+\lambda\!\sum_{a\in\{\operatorname{Re},\operatorname{Im}\}}
\rho\!\left(a\!\left(\operatorname{FFT}(\mathbf{x}_{0})\right),
a\!\left(\operatorname{FFT}(\hat{\mathbf{x}}_{0})\right)\right),
\end{split}
\end{equation}
where \(\rho\) is the configured pointwise loss (\(\ell_1\) in our experiments). This notation consistently compares the predicted and target clean sequences, rather than \(\mathbf{x}_s\).
UPLOTS conditions the reverse model on \(p_k\) through TPEM, as detailed below. The forward schedule and reverse coefficients are unchanged from Diffusion-TS. The prompt affects the step-specific estimate \(\hat{\mathbf{x}}_0\).
\paragraph{Interpretable decoder.} The network's output is factorised into trend, seasonality, and residual components, giving each generated sequence an explicit decomposition:
\begin{equation}\label{diff3}
        \hat{x}_{0}= 
        \underbrace{P(k)}_{\text{trend}} +
        \underbrace{\sum_{m=1}^{K}A_{m}\cos(2\pi f_{m}k+\phi_{m})}_{\text{seasonality}} + \underbrace{r(k)}_{\text{residual}}
\end{equation}

\section{Methodology}
\label{headings}

\begin{figure*}[!ht]
\centering
\includegraphics[width=0.99\linewidth]{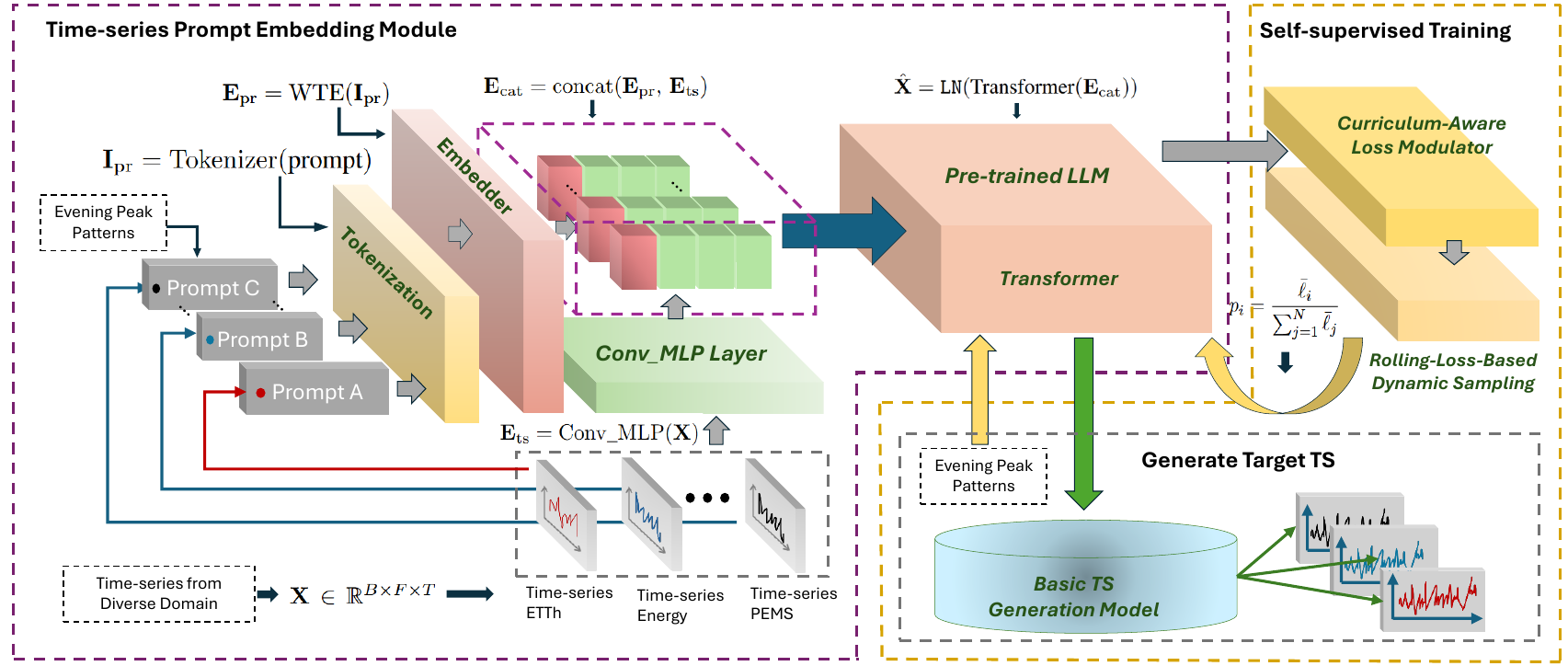}
\caption{Overview of UPLOTS. During training, channel-wise sequences and their dataset--scenario prompts are encoded by TPEM and passed to the Diffusion-TS encoder--decoder. CALM weights per-pool losses and RLDS adjusts per-pool micro-batch frequency. At inference, the prompt selects one of the learned scenario distributions.}
\Description{The UPLOTS pipeline concatenates prompt-token embeddings with projected channel-wise time-series inputs, processes the combined representation with a pretrained language model, and passes it to a time-series generator. CALM adjusts per-pool loss weights and RLDS adjusts pool sampling frequencies.}
\label{fig:over}
\end{figure*}

\subsection{Unified Pretrained LM Framework}
As shown in Figure~\ref{fig:over}, the conventional setup trains an individual generator for each dataset. UPLOTS instead shares one channel-independent generator across the selected dataset--scenario pools and conditions it on a pool identifier.

The shared representation is implemented with GPT-2~\cite{radford2019language} or LLaMA 3.2-1B~\cite{grattafiori2024llama}, followed by the trainable Diffusion-TS encoder and decoder. Prompt conditioning distinguishes the predefined pools, while the Dynamic Weighted Training Strategy addresses unequal loss scales and update requirements during joint training. Our empirical claim is limited to the datasets and prompt configurations evaluated in Section~\ref{exp}.

\subsection{Time-series Prompt Embedding Module}

In our unified framework, we jointly train one prompt-conditioned model on multiple dataset--scenario pools. Each pool is assigned a compact textual identifier, such as \texttt{ETTHMP}. At inference, the identifier selects the corresponding learned distribution.

The \textit{Time-series Prompt Embedding Module} (TPEM) concatenates token embeddings with projected time-series inputs before the truncated language-model blocks. Its role is conditioning: the experiment does not require the short prompt codes to express a compositional natural-language description.


Given a channel-wise input tensor \(\mathbf{X}\in\mathbb{R}^{B\times T\times 1}\), where \(B\) is the batch size and \(T\) the sequence length, a convolutional MLP maps it to the language-model width \(d_{\mathrm{LM}}\):

\begin{equation}\label{eq1}
  \mathbf{E}_{\mathrm{ts}} = \mathrm{ConvMLP}(\mathbf{X})\in\mathbb{R}^{B\times T\times d_{\mathrm{LM}}}.
\end{equation}

A textual identifier is tokenized and mapped through the language model's word-embedding table:


\begin{equation}
\mathbf{I}_{\mathrm{pr}} = \mathrm{Tokenizer}(\text{prompt}) \in \mathbb{Z}^{B \times L},
\end{equation}
where $B$ is the batch size and $L$ is the prompt length.
These IDs are then passed through the frozen word-embedding table of the pretrained LLM,

\begin{equation}
\mathbf{E}_{\mathrm{pr}} = \mathrm{WTE}(\mathbf{I}_{\mathrm{pr}}) \in \mathbb{R}^{B \times L \times d_{\mathrm{LM}}},
\end{equation}
where \(L\) is the tokenized prompt length. For the GPT-2 implementation, we retain the first two transformer blocks. The token embedding and attention/MLP weights in the GPT backbone are frozen. GPT LayerNorm and positional embeddings remain trainable. The input projection, output projections, padding token, Diffusion-TS encoder--decoder, and other surrounding layers are also trained. Thus, ``frozen backbone'' refers only to the core pretrained GPT parameters, not to the full UPLOTS model.

We concatenate the prompt and sequence embeddings along the temporal dimension:

\begin{equation}
 \mathbf{Z}_0= [\mathbf{E}_{\mathrm{pr}}; \mathbf{E}_{\mathrm{ts}}] \in \mathbb{R}^{B \times (L+T) \times d_{\mathrm{LM}}},
\label{eq:tpem_input_concat}
\end{equation}
where $[\cdot;\cdot]$ denotes concatenation along the temporal dimension.

The concatenated representation is processed by the truncated transformer. Two explicitly named linear projections then map the hidden width and padded token axis to the Diffusion-TS representation:

\begin{equation}\label{eq4}
\begin{split}
  \mathbf{H} &= \operatorname{Transformer}(\mathbf{Z}_0)
  \in \mathbb{R}^{B\times(L+T)\times d_{\mathrm{LM}}},\\
  \mathbf{E} &= \operatorname{Proj}_{\mathrm{token}}
  \!\left(\operatorname{Proj}_{\mathrm{hidden}}(\mathbf{H})\right)
  \in \mathbb{R}^{B\times T\times d_h}.
\end{split}
\end{equation}
Here \(\operatorname{Proj}_{\mathrm{hidden}}\) and \(\operatorname{Proj}_{\mathrm{token}}\) are linear layers and are distinct from LayerNorm.

\paragraph{Prompt-conditioned reverse diffusion.}
At diffusion step \(s\), the input \(\mathbf{X}\) in Equation~\ref{eq1} is the current noisy sample \(\mathbf{x}_s\). Although the selected prompt \(p_k\) stays fixed throughout sampling, \(\mathbf{x}_s\) changes, so the fused representation is recomputed at every reverse step:
\begin{equation}\label{eq:prompt_reverse}
\begin{aligned}
\mathbf{E}^{(k)}_s &= \operatorname{TPEM}(\mathbf{x}_s,p_k),\\
\mathbf{h}^{(k)}_s &= \operatorname{Enc}_{\theta}(\mathbf{E}^{(k)}_s,s),\\
\hat{\mathbf{x}}^{(s,k)}_0
&= \operatorname{Dec}_{\theta}(\mathbf{E}^{(k)}_s,\mathbf{h}^{(k)}_s,s),\\
\mathbf{x}_{s-1} &\sim p_{\theta}(\mathbf{x}_{s-1}\mid\mathbf{x}_s,p_k).
\end{aligned}
\end{equation}
The trend, seasonality, and residual outputs sum to \(\hat{\mathbf{x}}^{(s,k)}_0\), which enters the unchanged Diffusion-TS reverse update. Training uses the same conditioned path on randomly noised inputs with Equation~\ref{diff2}. Thus, the language model supplies prompt-aware representations. The Diffusion-TS encoder--decoder, rather than autoregressive language-model decoding, performs denoising and clean-sequence recovery.

\input{table/main1}

\subsection{Dynamic Weighted Training Strategy}

\subsubsection{Curriculum-Aware Loss Modulator}
\label{curr}
For each dataset--scenario pool \(i\), we keep a deque \(\mathcal{H}_i\) of the most recent \(W\) raw (unweighted) losses and compute
\begin{equation}\label{eq5}
\bar\ell_i=
\begin{cases}
\frac{1}{|\mathcal{H}_i|}\sum_{k\in\mathcal{H}_i}\ell_k,&|\mathcal{H}_i|>0,\\
1.0,&|\mathcal{H}_i|=0,
\end{cases}
\end{equation}
during training.  
Let \(\ell_{\max}=\max_i\bar\ell_i\) and \(\ell_{\min}=\min_i\bar\ell_i\).  
The normalized curriculum weight is
\begin{equation}\label{eq6}
w_i=
\begin{cases}
s+\dfrac{\ell_{\max}-\bar\ell_i}{\ell_{\max}-\ell_{\min}}(1-s),&\ell_{\max}>\ell_{\min},\\
1,&\ell_{\max}=\ell_{\min},
\end{cases}
\end{equation}
with \(s\in(0,1)\) denoting the minimum weight. Higher-loss pools receive smaller weights. We update these weights during the first \(E_w=5\) epochs, freeze the final warm-up weights for \(E_w\le e<E_a\), and set all weights to one after \(E_a=50\). The micro-batch loss is \(w_i\mathcal{L}_{\mathrm{diff}}\) while CALM is active.

\subsubsection{Rolling-Loss Dynamic Sampling}
Each pool's deque \(\mathcal{H}_i\) from Equation~\ref{eq5} is also used to derive micro-batch sampling probabilities
\begin{equation}\label{eq7}
p_i=\frac{\bar\ell_i^{1/\gamma}}{\sum_{j=1}^{K}\bar\ell_j^{1/\gamma}},
\end{equation}
where \(\gamma>0\) is a temperature (\(\gamma=1\) by default), so higher-loss pools are sampled more frequently. Each epoch retains a fixed micro-batch budget equal to the sum of the original loader lengths. Loaders are cyclic: if a selected loader is exhausted, its iterator is restarted, allowing RLDS to change the realized per-pool counts while keeping the total number of updates fixed. When RLDS is disabled, a shuffled static schedule uses each loader's original batch count exactly once.

The prompt identifies the target pool, CALM changes its loss contribution, and RLDS changes its update frequency. Their effects are evaluated separately in Section~\ref{abla}.

\section{Experiments}
\label{exp}

This section evaluates generation fidelity and downstream utility. We first compare dataset-specific baselines and multi-dataset methods, then analyze prompt configurations, model components, hyperparameters, and forecasting augmentation. Each generator configuration uses one training seed. The \(\pm\) values in the generation-quality tables are the standard deviations of five stochastic metric-estimation repeats on the same saved synthetic sample set. They are not variances over independently trained generators.



\begin{table}[h]
\centering
\normalsize
\caption{Source datasets. Scenario-pool definitions and sizes are reported in Appendix~\ref{app:scenario}.}
\label{tab:data}
\renewcommand\arraystretch{0.93}
\setlength{\tabcolsep}{0.8mm}
\begin{tabular}{lrrll}
\toprule
\textbf{Dataset} & \textbf{Obs.} & \textbf{Ch.} & \textbf{Period} & \textbf{Frequency} \\
\midrule
ETTh    & 17,420 & 7   & Jul.\ 2016--Jun.\ 2018 & 1 hour \\
Energy  & 19,735 & 28  & Jan.--May 2016 & 10 minutes \\
PEMS04  & 16,992 & 307 & Jan.--Feb.\ 2018 & 5 minutes \\
PEMS08  & 17,856 & 170 & Jul.--Aug.\ 2016 & 5 minutes \\
\bottomrule
\end{tabular}
\end{table}

\subsection{Experimental Settings}

\paragraph{Datasets and scenario protocol} We use the four timestamped datasets in Table~\ref{tab:data}. For the main comparison, observations are assigned to morning-peak (07:00--10:00) or evening-peak (17:00--20:00) reference pools before sliding-window extraction. The extended ETTh/Energy study additionally forms workday/weekend, high/low-load, and volatile pools. Appendix~\ref{app:scenario} gives the exact rules and sizes. This construction defines a \emph{scenario-conditioned subset-generation} task: a prompt selects the distribution of a predefined, seen pool. It should not be interpreted as editing a localized interval inside a contiguous full-day trajectory, composing unseen constraints, or generating an unseen domain. Consistent with common in-distribution TSG evaluation, synthetic samples are compared with real sequences from the corresponding scenario reference pool. We therefore do not describe the main tables as held-out or zero-shot generalization.

\paragraph{Channel-independent evaluation} Each multivariate segment is split into channel-wise univariate sequences and processed by the same generator, allowing parameters to be shared across datasets with different channel counts. The reported generation metrics use the corresponding pooled channel-wise real and synthetic samples.

\paragraph{Metrics}
We use the following three metrics as primary measures of distributional fidelity, temporal structure, and downstream predictive utility:
\begin{itemize}
\item \textbf{Discriminative Score}~\cite{yoon2019time} evaluates distributional fidelity by training a binary classifier to distinguish real from generated time-series. 
\item \textbf{Predictive Score}~\cite{yoon2019time} measures predictive utility under the Train-on-Synthetic, Test-on-Real (TSTR) paradigm, whereby a sequence forecasting model is trained on synthetic data and evaluated on real observations using next-step prediction error. 
\item \textbf{Context-FID} (Context--Fréchet Inception Distance)~\cite{jehapsa} compares local-subsequence embeddings from pooled real and synthetic channel sequences.
\end{itemize}
These metrics do not directly test whether generated samples are closer to the requested scenario than to an alternative. As a supplementary post-hoc diagnostic, Appendix~\ref{app:sps} reports a target-versus-contrast Scenario Preference Score (SPS). SPS is computed from generated samples and is not used for training, hyperparameter selection, model selection, or the main method ranking.

\begin{figure}[t!]
  \centering
  \includegraphics[width=\linewidth]{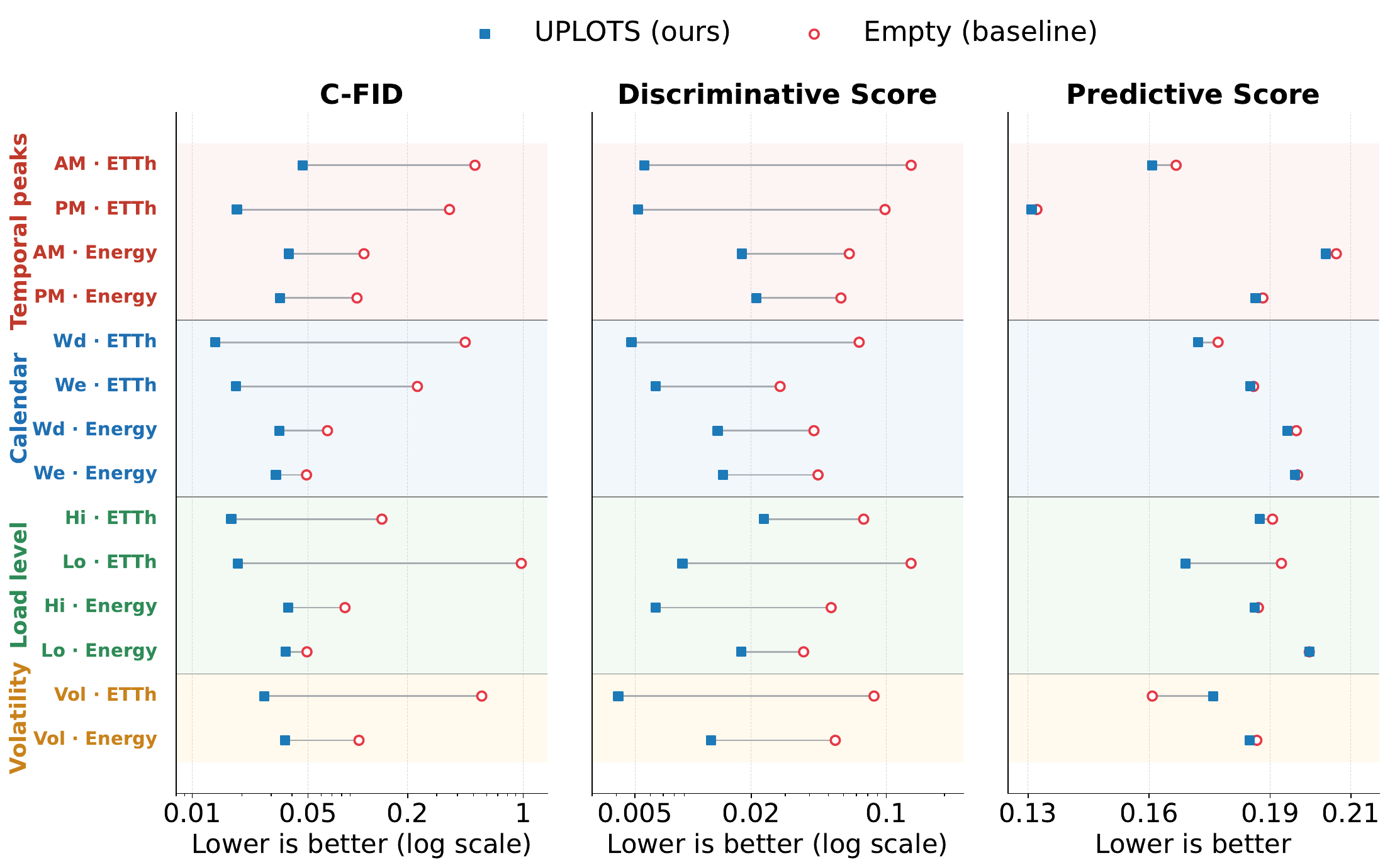}
  \caption{%
    \textbf{Extended evaluation across 14 prompt configurations} on ETTh and Energy, spanning four constraint families (\textit{temporal peaks}, \textit{calendar}, \textit{load level}, \textit{volatility}). \textcolor{myblue}{$\blacksquare$}~\textsc{UPLOTS} (ours) vs.\ \textcolor{myred}{$\bigcirc$}~\textsc{Empty} baseline; lower is better in all three panels (C-FID and Discriminative Score on logarithmic scales).%
  }
  \Description{Three dot plots compare UPLOTS with the Empty baseline using Context-FID, Discriminative Score, and Predictive Score for 14 ETTh and Energy prompt configurations grouped by temporal peak, calendar, load level, and volatility.}
  \label{fig_all14}
\end{figure}

\subsection{Main Results}
Table~\ref{tab:exp} compares five dataset-specific baselines: TimeGAN~\cite{yoon2019time}, TimeVAE~\cite{desai2021timevae}, DiffTime~\cite{coletta2023constrained}, DiffWave~\cite{kongdiffwave}, and Diffusion-TS~\cite{yuandiffusion}. It also reports multi-dataset methods. TimeDP~\cite{huang2025timedp} is the closest multi-domain baseline, although it derives prototype prompts from few-shot target examples, whereas UPLOTS uses predefined text identifiers. We compare (i) UPLOTS (Empty), without TPEM or the dynamic weighting strategy, (ii) UPLOTS (GPT-2), and (iii) UPLOTS (LLaMA). Under this in-distribution protocol, the two prompt-conditioned variants substantially reduce Context-FID relative to Empty and are competitive with or better than the listed baselines on most dataset--scenario pairs. Differences between GPT-2 and LLaMA are small and mixed, so the table does not by itself establish that a larger language model is necessary. Each row reports one training run. The \(\pm\) values are the standard deviations of five stochastic evaluator repeats as defined above.

Figure~\ref{fig_all14} broadens the in-distribution evaluation to 14 prompt configurations spanning temporal peaks, calendar, load level, and volatility on ETTh and Energy, and visualizes the three primary metrics for UPLOTS and Empty. These metrics do not alone establish scenario preference. The supplementary target-versus-contrast analysis in Appendix~\ref{app:sps} shows dataset-dependent behavior, including weaker preference for several Energy scenarios, and is not part of the headline method ranking.

\begin{table}[!h]
\centering
\small
\caption{Length-24 ablations over four seen peak pools. C-FID is the four-pool macro-average. SPS is a supplementary post-hoc diagnostic in percent and is not used to select variants. Each variant uses one training seed. C-FID uses three evaluator fits per pool, and SPS uses 100 sample-level bootstrap resamples. Bold marks the best point estimate per column.}
\label{ablation}
\renewcommand{\arraystretch}{0.94}
\setlength{\tabcolsep}{1.0mm}
\begin{tabular}{lrrrr}
\toprule
\textbf{Variant} & \textbf{C-FID} & \multicolumn{3}{c}{\textbf{SPS (\%)}}\\
\cmidrule(lr){3-5}
 & \textbf{$\downarrow$} & \textbf{ETTh} & \textbf{Energy} & \textbf{Mean}\\
\midrule
Pretr., text, frozen (full) & \textbf{0.0284} & 81.39 & \textbf{53.36} & 67.37\\
Random, text, frozen & 0.0315 & 73.57 & 53.07 & 63.32\\
Random, text, trainable & 0.0352 & 71.27 & 52.39 & 61.83\\
Pretr., text, trainable & 0.0298 & \textbf{82.67} & 52.97 & 67.82\\
Pretr., learned ID, frozen & 0.0292 & 82.29 & 53.26 & 67.77\\
Pretr., no prompt, frozen & 0.1124 & 50.05 & 49.92 & 49.98\\
\midrule
Full w/o CALM & 0.0289 & 82.36 & 53.32 & \textbf{67.84}\\
Full w/o RLDS & 0.0300 & 81.70 & 52.89 & 67.29\\
\bottomrule
\end{tabular}
\end{table}

\subsection{Ablation Study}\label{abla}

Table~\ref{ablation} separates backbone initialization and training, conditioning form, and the effects of CALM and RLDS. The full configuration uses pretrained text embeddings, a frozen GPT core, CALM, and RLDS.

Removing only the prompt tokens changes C-FID from 0.0284 to 0.1124, showing the value of scenario conditioning in this one-seed, four-pool run. For reference, the supplementary SPS diagnostic changes from 67.37\% to 49.98\%. Text and learned IDs yield similar point estimates (C-FID 0.0284 versus 0.0292; SPS 67.37\% versus 67.77\%), so we claim no unique semantic benefit for text. Pretraining gives better point estimates than random initialization in both matched frozen and trainable comparisons. Freezing and the training strategies show metric-dependent trade-offs. These descriptive results do not establish statistical superiority or necessity.

\begin{figure}[!h]
\centering
\includegraphics[width=0.87\linewidth]{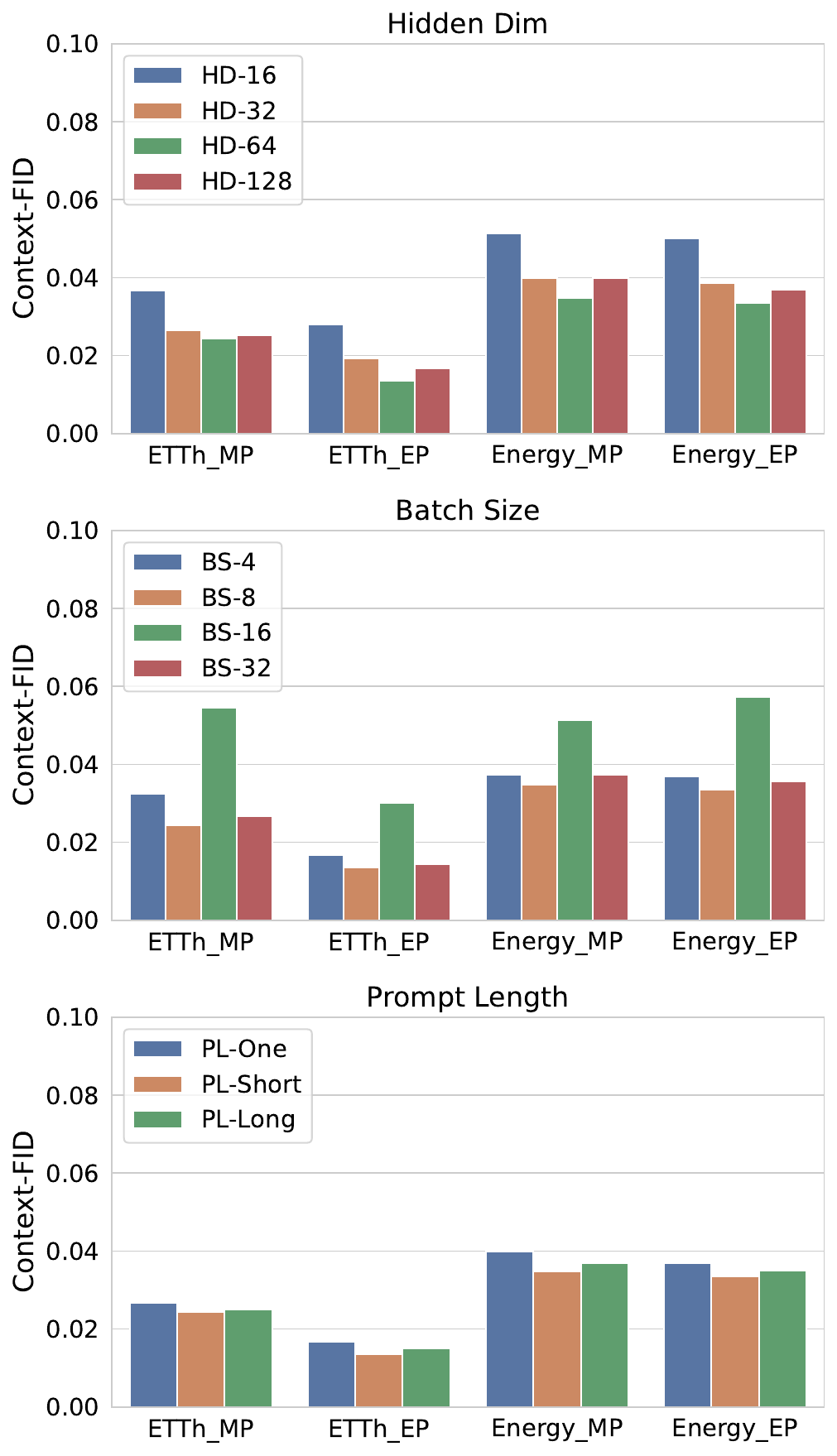}
\caption{Effect of hyperparameters on Context-FID.}
\Description{Three grouped bar charts show Context-FID across tested hidden dimensions, batch sizes, and prompt lengths for ETTh and Energy morning- and evening-peak pools.}
\label{hyper}
\end{figure}

\subsection{Downstream Task}

We train PhaseFormer~\cite{niu2025phaseformer} and SparseTSF~\cite{lin2024sparsetsf} on morning/evening peak segments of ETTh and Energy (12\(\rightarrow\)12), varying the real-data ratio from 1\% to 90\%, and compare real-only training with real plus UPLOTS-generated data.
Table~\ref{tab:aug_results} shows that augmentation is most helpful in the low-data regimes. At high real-data ratios, several SparseTSF entries are tied or slightly worse, so we interpret the experiment as evidence of utility under scarcity rather than a universal forecasting improvement.

\begin{table}[!h]
\centering
\caption{Performance comparison between original training data (\textit{orig}) and augmented training data (\textit{aug}) under different data ratios. Lower is better for both MSE and MAE.}
\label{tab:aug_results}
\scriptsize
\renewcommand{\arraystretch}{1.0}
\setlength{\tabcolsep}{0.75pt}

\begin{subtable}{\linewidth}
\centering
\subcaption{PhaseFormer}
\begin{tabular}{@{}llccccc@{}}
\toprule
\multirow{2}{*}{\textbf{Dataset}} & \multirow{2}{*}{\textbf{Metric}}
& \multicolumn{5}{c}{\textbf{Real-data Ratio} (\textit{orig}/\textit{aug})} \\
\cmidrule(lr){3-7}
 &  & \textbf{1\%} & \textbf{5\%} & \textbf{10\%} & \textbf{30\%} & \textbf{90\%} \\
\midrule
\multirow{2}{*}{ETTh\_MP}   & MSE & 0.3439 / \textbf{0.2804} & 0.3043 / \textbf{0.2805} & 0.2995 / \textbf{0.2793} & 0.2909 / \textbf{0.2783} & 0.2836 / \textbf{0.2745} \\
                            & MAE & 0.4066 / \textbf{0.3682} & 0.3778 / \textbf{0.3682} & 0.3745 / \textbf{0.3680} & 0.3717 / \textbf{0.3678} & 0.3695 / \textbf{0.3651} \\
\cmidrule(lr){1-7}
\multirow{2}{*}{ETTh\_EP}   & MSE & 0.2667 / \textbf{0.2147} & 0.2380 / \textbf{0.2144} & 0.2315 / \textbf{0.2141} & 0.2214 / \textbf{0.2134} & 0.2174 / \textbf{0.2116} \\
                            & MAE & 0.3419 / \textbf{0.3032} & 0.3207 / \textbf{0.3028} & 0.3145 / \textbf{0.3026} & 0.3077 / \textbf{0.3019} & 0.3053 / \textbf{0.3005} \\
\cmidrule(lr){1-7}
\multirow{2}{*}{Energy\_MP} & MSE & 0.2963 / \textbf{0.2700} & 0.2868 / \textbf{0.2699} & 0.2832 / \textbf{0.2698} & 0.2764 / \textbf{0.2671} & 0.2661 / \textbf{0.2638} \\
                            & MAE & 0.3306 / \textbf{0.3010} & 0.3196 / \textbf{0.3019} & 0.3173 / \textbf{0.3031} & 0.3124 / \textbf{0.2985} & 0.3029 / \textbf{0.2978} \\
\cmidrule(lr){1-7}
\multirow{2}{*}{Energy\_EP} & MSE & 0.3388 / \textbf{0.3060} & 0.3262 / \textbf{0.3063} & 0.3202 / \textbf{0.3049} & 0.3116 / \textbf{0.3029} & 0.3030 / \textbf{0.2983} \\
                            & MAE & 0.3671 / \textbf{0.3305} & 0.3520 / \textbf{0.3296} & 0.3480 / \textbf{0.3294} & 0.3416 / \textbf{0.3261} & 0.3326 / \textbf{0.3235} \\
\bottomrule
\end{tabular}
\end{subtable}


\begin{subtable}{\linewidth}
\centering
\subcaption{SparseTSF}
\begin{tabular}{@{}llccccc@{}}
\toprule
\multirow{2}{*}{\textbf{Dataset}} & \multirow{2}{*}{\textbf{Metric}}
& \multicolumn{5}{c}{\textbf{Real-data Ratio} (\textit{orig}/\textit{aug})} \\
\cmidrule(lr){3-7}
 &  & \textbf{1\%} & \textbf{5\%} & \textbf{10\%} & \textbf{30\%} & \textbf{90\%} \\
\midrule
\multirow{2}{*}{ETTh\_MP}   & MSE & 0.4840 / \textbf{0.2899} & 0.3861 / \textbf{0.2899} & 0.3356 / \textbf{0.2898} & 0.2969 / \textbf{0.2898} & 0.2896 / 0.2898 \\
                            & MAE & 0.4855 / \textbf{0.3705} & 0.4336 / \textbf{0.3705} & 0.4021 / \textbf{0.3705} & 0.3752 / \textbf{0.3704} & 0.3703 / 0.3705 \\
\cmidrule(lr){1-7}
\multirow{2}{*}{ETTh\_EP}   & MSE & 0.3671 / \textbf{0.2208} & 0.3024 / \textbf{0.2208} & 0.2663 / \textbf{0.2207} & 0.2350 / \textbf{0.2205} & 0.2208 / \textbf{0.2206} \\
                            & MAE & 0.4055 / \textbf{0.3087} & 0.3662 / \textbf{0.3087} & 0.3414 / \textbf{0.3086} & 0.3188 / \textbf{0.3086} & 0.3084 / 0.3085 \\
\cmidrule(lr){1-7}
\multirow{2}{*}{Energy\_MP} & MSE & 0.3792 / \textbf{0.2832} & 0.3045 / \textbf{0.2837} & 0.2914 / \textbf{0.2835} & 0.2831 / \textbf{0.2831} & 0.2814 / 0.2822 \\
                            & MAE & 0.3775 / \textbf{0.3157} & 0.3455 / \textbf{0.3151} & 0.3350 / \textbf{0.3152} & 0.3202 / \textbf{0.3155} & 0.3187 / \textbf{0.3166} \\
\cmidrule(lr){1-7}
\multirow{2}{*}{Energy\_EP} & MSE & 0.4305 / \textbf{0.3201} & 0.3467 / \textbf{0.3202} & 0.3316 / \textbf{0.3198} & 0.3208 / \textbf{0.3197} & 0.3186 / 0.3190 \\
                            & MAE & 0.4144 / \textbf{0.3468} & 0.3808 / \textbf{0.3462} & 0.3704 / \textbf{0.3466} & 0.3548 / \textbf{0.3465} & 0.3509 / \textbf{0.3483} \\
\bottomrule
\end{tabular}
\end{subtable}

\end{table}

\subsection{Hyperparameter Analysis}
Figure~\ref{hyper} reports the tested hidden dimensions, batch sizes, and TPEM prompt lengths using Context-FID. Within this grid, hidden dimension 64 and batch size 32 provide a favorable setting, while the prompt-length variants differ modestly.

We define PL-One as a single-character identifier (e.g., \texttt{A}), PL-Short as a multi-letter code (e.g., \texttt{ETTHMP}), and PL-Long as a phrase (e.g., \emph{The morning peak patterns in the ETTh dataset}). Because these variants change both token identity and length, this study does not isolate semantic understanding. We use hidden dimension 64, batch size 32, and PL-Short for the main results.

\subsection{Visualization and Analysis}

\begin{figure*}[!ht]
  \centering
  \begin{subfigure}[b]{0.91\linewidth}
    \centering
    \includegraphics[width=\linewidth]{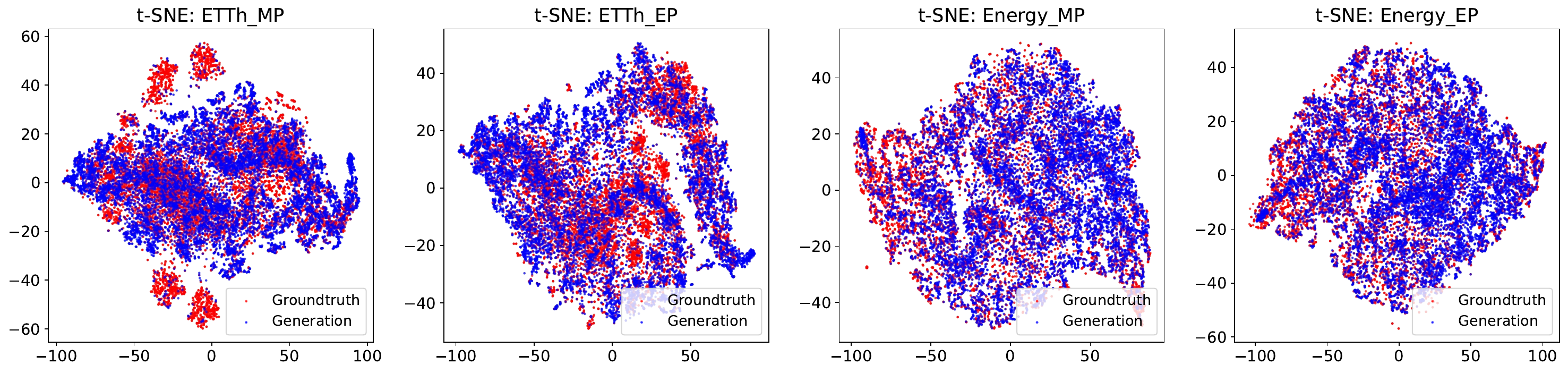}
    \caption{Diffusion-TS}
    \label{tsne1}
  \end{subfigure}\\[1ex]
  \begin{subfigure}[b]{0.91\linewidth}
    \centering
    \includegraphics[width=\linewidth]{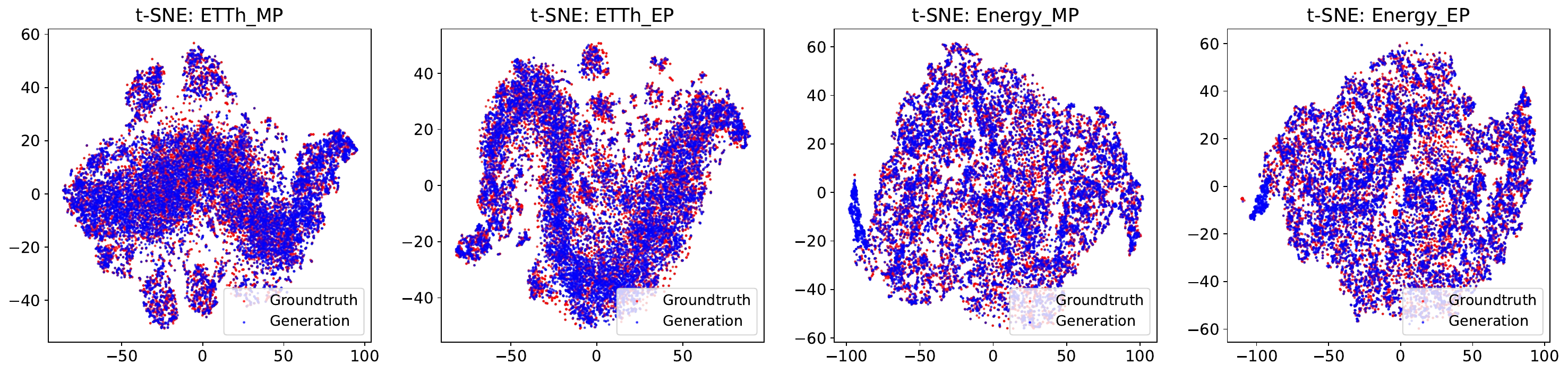}
    \caption{UPLOTS}
    \label{tsne2}
  \end{subfigure}
  \caption{Visualizations of the time-series synthesized by (a) Diffusion-TS and (b) UPLOTS.}
  \Description{Eight t-SNE scatter plots compare real and generated channel-wise sequences for ETTh and Energy morning- and evening-peak pools. The upper four panels show Diffusion-TS and the lower four show UPLOTS.}
  \label{tsne}
\end{figure*}

\begin{figure*}[!ht]
  \centering
  \begin{subfigure}[b]{0.27\linewidth}
    \centering
    \includegraphics[width=\linewidth]{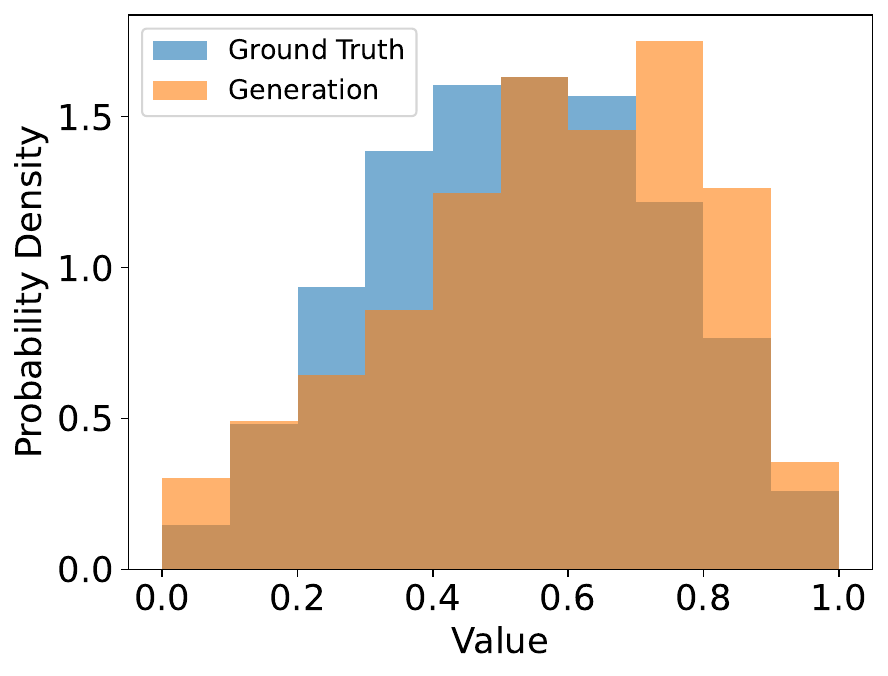}
    \caption{Diffusion-TS}
    \label{exp:sub1}
  \end{subfigure}
  \hfill
  \begin{subfigure}[b]{0.27\linewidth}
    \centering
    \includegraphics[width=\linewidth]{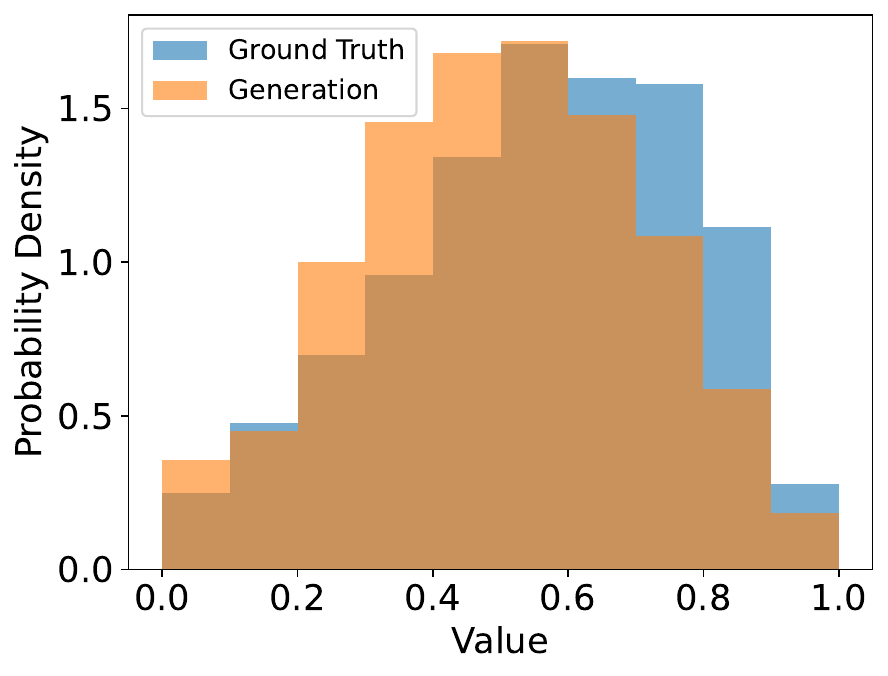}
    \caption{UPLOTS (Empty)}
    \label{exp:sub2}
  \end{subfigure}
    \hfill
  \begin{subfigure}[b]{0.27\linewidth}
    \centering
    \includegraphics[width=\linewidth]{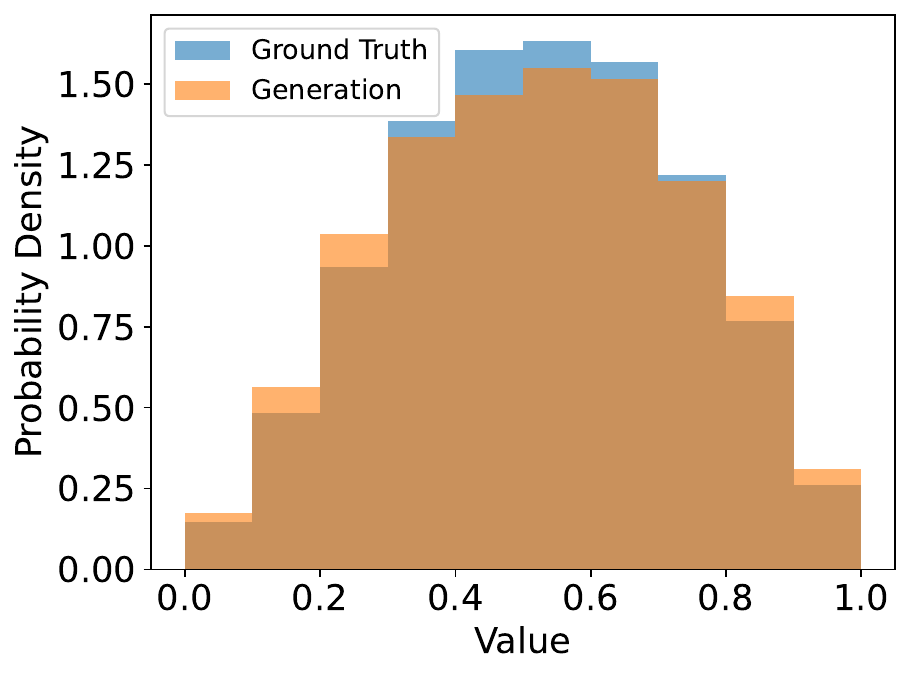}
    \caption{UPLOTS}
    \label{exp:sub3}
  \end{subfigure}  
  \caption{Marginal density estimates for Diffusion-TS, UPLOTS (Empty), and the full UPLOTS model.}
    \Description{Three panels of overlaid histograms compare ground-truth and generated value distributions for Diffusion-TS, UPLOTS (Empty), and the full UPLOTS model.}
  \label{exp2}
\end{figure*}

Figure~\ref{tsne} provides a qualitative t\hyp{}SNE projection of real and synthetic channel-wise sequences for \textit{ETTh\_MP/EP} and \textit{Energy\_MP/EP}. UPLOTS samples overlap the real point clouds more closely than Diffusion-TS in these two-dimensional projections. Because t\hyp{}SNE is stochastic and distorts global distances, we use this figure as supporting visualization rather than quantitative evidence of manifold recovery.

We also compare marginal density estimates in Figure~\ref{exp2}. The UPLOTS curve is closer to the real-data curve than the displayed alternatives over most of the plotted range, complementing the quantitative fidelity metrics.

\section{Conclusion}
This paper introduced UPLOTS, a shared prompt-guided framework for channel-independent time-series generation over multiple predefined dataset--scenario pools. A compact textual identifier conditions a mostly frozen language-model backbone inside a trainable Diffusion-TS generator, while CALM and RLDS balance joint training.

Experiments on four benchmarks show improved in-distribution generation fidelity for the morning/evening pools, and the extended ETTh/Energy study covers 14 prompt configurations. Forecasting augmentation is most useful when real training data are scarce.

In the one-seed ablation, removing prompt tokens produces worse point estimates, whereas text and learned IDs perform similarly. The observed effects of pretraining, freezing, CALM, and RLDS depend on the metric. They do not establish statistical advantages.

We do not evaluate unseen domains or new prompt combinations. Future work should build continuity-preserving constraint benchmarks, add held-out evaluations, and model cross-channel dependencies explicitly.



\appendix
\section{Appendix}
\label{appendix}

\subsection{Scenario-Pool Construction}
\label{app:scenario}
The peak pools contain observations whose timestamps fall in 07:00--10:00 (MP) or 17:00--20:00 (EP), inclusive. The calendar split uses the workday/holiday indicator of the preprocessing calendar, with its complement forming WKE. For load level, the primary signal is OT for ETTh, Appliances for Energy, and mean sensor flow for PEMS. HI and LO retain observations at or above the 75th percentile and at or below the 25th percentile, respectively. VOL retains observations whose absolute first difference in the same primary signal is at or above its 75th percentile. After filtering, observations remain in their original order and sliding windows are extracted from each resulting pool. Consequently, these are scenario-specific subset distributions rather than localized edits to contiguous full-day trajectories.

\begin{table}[H]
\centering
\scriptsize
\caption{Numbers of observations in the scenario pools before sliding-window extraction. MP: morning peak; EP: evening peak; WKD/WKE: workday/non-workday; HI/LO: high/low load; VOL: volatile.}
\label{tab:scenario_sizes}
\setlength{\tabcolsep}{1.8mm}
\begin{tabular}{lrrrrrrr}
\toprule
\textbf{Dataset} & \textbf{MP} & \textbf{EP} & \textbf{WKD} & \textbf{WKE} & \textbf{HI} & \textbf{LO} & \textbf{VOL}\\
\midrule
ETTh   & 2,904 & 2,903 & 11,900 & 5,520 & 4,369 & 4,356 & 4,370\\
Energy & 2,603 & 2,610 & 13,543 & 6,192 & 5,211 & 7,462 & 6,355\\
PEMS04 & 2,183 & 2,183 & 11,232 & 5,760 & 4,248 & 4,248 & 4,248\\
PEMS08 & 2,294 & 2,294 & 12,672 & 5,184 & 4,464 & 4,464 & 4,464\\
\bottomrule
\end{tabular}
\end{table}

\begin{table}[H]
\centering
\small
\caption{Individual SPS results in percent with 95\% sample-level bootstrap intervals. Values above 50 indicate that generated samples are closer to the requested scenario than to the listed contrast.}
\label{tab:sps_full}
\setlength{\tabcolsep}{2.0mm}
\begin{tabular}{llll}
\toprule
\textbf{Dataset} & \textbf{Prompt} & \textbf{Contrast} & \textbf{SPS [95\% CI]}\\
\midrule
ETTh & Morning peak & Evening peak & 59.3 [52.5, 64.6]\\
ETTh & Evening peak & Morning peak & 75.2 [72.0, 78.4]\\
ETTh & Workday & Weekend & 66.6 [55.8, 75.1]\\
ETTh & Weekend & Workday & 77.2 [71.8, 82.0]\\
ETTh & High load & Low load & 89.4 [85.8, 92.1]\\
ETTh & Low load & High load & 92.7 [90.5, 94.9]\\
ETTh & Volatile & Non-volatile pool & 89.0 [86.1, 91.5]\\
\midrule
Energy & Morning peak & Evening peak & 54.3 [47.5, 61.9]\\
Energy & Evening peak & Morning peak & 52.2 [43.9, 60.2]\\
Energy & Workday & Weekend & 52.9 [45.3, 60.8]\\
Energy & Weekend & Workday & 50.5 [43.9, 58.8]\\
Energy & High load & Low load & 59.1 [53.7, 66.5]\\
Energy & Low load & High load & 57.8 [52.2, 63.1]\\
Energy & Volatile & Non-volatile pool & 51.5 [43.0, 60.2]\\
\bottomrule
\end{tabular}
\end{table}

\subsection{Scenario Preference Details}
\label{app:sps}
The standard metrics in the main text remain our primary evaluation measures. We report the Scenario Preference Score (SPS) only as a supplementary post-hoc diagnostic of whether generated samples are closer to a requested scenario than to a paired alternative. It is not used for training, hyperparameter selection, model selection, or the main method ranking. For generated samples \(G_c\), target real pool \(R_c\), and a paired contrast \(R_{\bar c}\), let
\[
d_c=\operatorname{FD}(\phi(G_c),\phi(R_c)),\qquad
d_{\bar c}=\operatorname{FD}(\phi(G_c),\phi(R_{\bar c})),
\]
where \(\phi\) contains transparent sequence summaries: mean, standard deviation, extrema, five quantiles, mean absolute first difference, first/second-half means, and lag-1/2 autocorrelations. We define
\begin{equation}\label{eq:sps}
\operatorname{SPS}(c)=\frac{d_{\bar c}}{d_c+d_{\bar c}}.
\end{equation}

An SPS above \(50\%\) means that generated samples are closer to the requested pool than to its contrast. We pair morning/evening, workday/weekend, and high/low load. The volatility contrast is the pooled non-volatile scenarios.

Table~\ref{tab:sps_full} reports the individual scores. The evaluation uses length-24 GPT-2 samples, pools at most 10,000 channel-wise sequences, and reports percentile intervals from 100 sample-level bootstrap resamples of size 3,000. These intervals quantify sampling/evaluation uncertainty for a fixed trained generator and must not be interpreted as uncertainty across training seeds. The ETTh results show clearer scenario preference, particularly for load and volatility. Energy is closer to the neutral boundary: the high/low-load intervals are above \(50\%\), whereas the temporal, calendar, and volatility intervals include \(50\%\). We therefore treat this diagnostic as dataset-dependent rather than claiming uniformly precise compliance.

\input{table/main48}

\subsection{Results}
To complement the main results, the Appendix reports additional quantitative comparisons under alternative generation lengths.
\input{table/main96}


Tables~\ref{tab:main48} and~\ref{tab:main48_pems} report peak-pool generation at length 48, while Tables~\ref{tab:main96} and~\ref{tab:main96_pems} report the corresponding results at length 96.

\FloatBarrier

\begin{acks}
This work was supported by the ARC Centre of Excellence for Automated Decision-Making and Society (CE200100005). We acknowledge the resources and services provided by the National Computational Infrastructure (NCI), which is supported by the Australian Government. This research was also partially supported by the ARC Training Centre for Whole Life Design of Carbon Neutral Infrastructure (IC230100015).
\end{acks}

\section*{GenAI Usage Disclosure}
The authors used Gemini and ChatGPT solely to assist with code review and language polishing, including identifying potential implementation inconsistencies and improving grammar, punctuation, clarity, and readability. All AI-assisted suggestions were reviewed and verified by the authors, who take full responsibility for the code, analyses, and final manuscript.

\bibliographystyle{ACM-Reference-Format}
\balance
\bibliography{acm}


\end{document}

%% file: table/main1.tex
\begin{table*}[h!]
\centering
\footnotesize
\caption{Comparison with baselines under dataset-specific and multi-dataset training. Lower is better. Each generator uses one training seed; \(\pm\) is the standard deviation of five stochastic evaluator repeats on the same generated sample set. Boldface and underlining mark the best value in each column.}
\label{tab:exp}
\renewcommand{\arraystretch}{0.97}
\setlength{\tabcolsep}{0.9mm}
\begin{tabular}{c c l c c c c c ccc}
\toprule
\hline
\textbf{Metric} & \textbf{Data Type} & \textbf{Model} & \textbf{ETTh\_MP} & \textbf{ETTh\_EP} & \textbf{Energy\_MP} & \textbf{Energy\_EP}  & \textbf{PEMS04\_MP} & \textbf{PEMS04\_EP} & \textbf{PEMS08\_MP} & \textbf{PEMS08\_EP}  \\
\hline
\multirow{9}{*}{\textbf{Context-FID}} 
   & \multirow{5}{*}{\textbf{\shortstack{Single\\Data}}} & TimeGAN-NIPS2019 & 0.2939\textpm.024 & 0.1365\textpm.019 & 0.2831\textpm.033 & 0.2512\textpm.029 & 0.0959\textpm.017 &  0.0619\textpm.018 & 0.0741\textpm.016 & 0.4383\textpm.054  \\
\cline{3-11}
& & TimeVAE-arXiv2022 & 0.6352\textpm.074 & 0.2634\textpm.046 & 0.3925\textpm.036 & 0.4712\textpm.044 & 0.4751\textpm.051 &  0.7330\textpm.084 & 0.3768\textpm.026 & 0.4882\textpm.033 \\
\cline{3-11}
& & DiffWave-ICLR2021 & 0.1613\textpm.016 & 0.1081\textpm.014 & 0.1125\textpm.011 & 0.1354\textpm.019 & 0.0874\textpm.009 & 0.0995\textpm.008 & 0.8951\textpm.012 & 0.9541\textpm.077 \\
\cline{3-11}
& & DiffTime-NIPS2023 & 0.1199\textpm.009 & 0.0652\textpm.008 & 0.0439\textpm.003 & 0.0526\textpm.007 & 0.0418\textpm.008 & 0.0866\textpm.013 & 0.0700\textpm.011 & 0.0757\textpm.009 \\
\cline{3-11}
& & Diffusion-TS-ICLR2024 & 0.0478\textpm.005 & 0.0153\textpm.002 & 0.0462\textpm.007 & 0.0419\textpm.003 & 0.0275\textpm.005 &  0.0139\textpm.003 & 0.0252\textpm.004 & 0.5037\textpm.081 \\
\cline{2-11}
   & \multirow{4}{*}{\textbf{\shortstack{Multi\\Data}}} & TimeDP-AAAI2025  & 0.2145\textpm.058 & 0.0951\textpm.024 & 0.0840\textpm.024 & 0.0856\textpm.025 & 0.1271\textpm.012 & 0.1417\textpm.012 & 0.1176\textpm.026 & 0.2340\textpm.054 \\
   \cline{3-11}
   & & UPLOTS (Empty) & 0.1682\textpm.031 & 0.1997\textpm.016 & 0.2908\textpm.025 & 0.2040\textpm.017 & 0.3285\textpm.027 & 0.6286\textpm.074 & 0.3276\textpm.051 & 0.5256\textpm.053 \\
   \cline{3-11}   
   & & UPLOTS (GPT-2) & 0.0243\textpm.002 & \cellcolor{blue!10} \textbf{\underline{0.0136\textpm.001}} & 0.0348\textpm.004 & 0.0334\textpm.002 & 0.0161\textpm.002 & 0.0073\textpm.001 & \cellcolor{blue!10} \textbf{\underline{0.0137\textpm.001}} & \cellcolor{blue!10} \textbf{\underline{0.0091\textpm.000}} \\
   \cline{3-11}
   & & UPLOTS (LLaMA) & \cellcolor{blue!10} \textbf{\underline{0.0226\textpm.002}} & 0.0145\textpm.002 & \cellcolor{blue!10} \textbf{\underline{0.0323\textpm.002}} & \cellcolor{blue!10} \textbf{\underline{0.0315\textpm.002}} & \cellcolor{blue!10} \textbf{\underline{0.0146\textpm.002}} & \cellcolor{blue!10} \textbf{\underline{0.0064\textpm.001}} & 0.0141\textpm.001 & 0.0093\textpm.000  \\
   \cline{3-11}
 
\hline
\hline
\multirow{9}{*}{\shortstack{\textbf{Discriminative}\\\textbf{Score}}}
   & \multirow{5}{*}{\textbf{\shortstack{Single\\Data}}} & TimeGAN-NIPS2019 & 0.0299\textpm.016 & 0.0151\textpm.009 & 0.0112\textpm.009 & 0.0605\textpm.007 & 0.0188\textpm.016 & 0.0402\textpm.024 & 0.0126\textpm.002 & 0.0270\textpm.024 \\
\cline{3-11}
& & TimeVAE-arXiv2022  & 0.0232\textpm.016 & 0.0135\textpm.011 & 0.0223\textpm.012 & 0.0303\textpm.007 & 0.0188\textpm.016 & 0.0716\textpm.054 & 0.0154\textpm.009 & 0.0453\textpm.016 \\
\cline{3-11}
& & DiffWave-ICLR2021 & 0.0158\textpm.010 & 0.0161\textpm.013 & 0.0194\textpm.009 & 0.0157\textpm.011 & 0.0236\textpm.021 & 0.0187\textpm.014 & 0.0249\textpm.006 & 0.0104\textpm.009 \\
\cline{3-11}
& & DiffTime-NIPS2023 & 0.0119\textpm.008 & 0.0140\textpm.010 & 0.0172\textpm.014 & 0.0175\textpm.013 & 0.0199\textpm.019 & 0.0164\textpm.011 & 0.0233\textpm.005 & 0.0080\textpm.006  \\
\cline{3-11}
& & Diffusion-TS-ICLR2024 & 0.0123\textpm.008 & 0.0122\textpm.009 & 0.0115\textpm.006 & 0.0169\textpm.011 & 0.0193\textpm.017 &  0.0063\textpm.009 & 0.0131\textpm.002 & 0.0711\textpm.019 \\
\cline{2-11}
   & \multirow{4}{*}{\textbf{\shortstack{Multi\\Data}}} & TimeDP-AAAI2025  & 0.0509\textpm.004 & 0.0300\textpm.010 & 0.0319\textpm.009 & 0.0349\textpm.009 & 0.0228\textpm.007 & 0.0239\textpm.013 & 0.0231\textpm.000 & 0.0194\textpm.011 \\
   \cline{3-11}
   &   & UPLOTS (Empty) & 0.0560\textpm.029 & 0.0950\textpm.009 & 0.0889\textpm.008 & 0.0833\textpm.006 & 0.1233\textpm.033 & 0.1210\textpm.074 & 0.0919\textpm.016 & 0.1670\textpm.033 \\
   \cline{3-11}
   & & UPLOTS (GPT-2) & 0.0066\textpm.004 & \cellcolor{blue!10} \textbf{\underline{0.0099}}\textpm.004 & 0.0077\textpm.004 & 0.0128\textpm.005 & 0.0181\textpm.014 & 0.0059\textpm.009 & 0.0129\textpm.001 & 0.0064\textpm.004 \\
   \cline{3-11}
   & & UPLOTS (LLaMA) & \cellcolor{blue!10} \textbf{\underline{0.0058}}\textpm.004  & 0.0103\textpm.007 & \cellcolor{blue!10} \textbf{\underline{0.0071}}\textpm.004 & \cellcolor{blue!10} \textbf{\underline{0.0116}}\textpm.004 & \cellcolor{blue!10} \textbf{\underline{0.0160}}\textpm.012 & \cellcolor{blue!10} \textbf{\underline{0.0048}}\textpm.008 & \cellcolor{blue!10} \textbf{\underline{0.0121}}\textpm.000 & \cellcolor{blue!10} \textbf{\underline{0.0059}}\textpm.004  \\
   \cline{3-11}

\hline
\multirow{9}{*}{\shortstack{\textbf{Predictive}\\\textbf{Score}}}
   & \multirow{5}{*}{\textbf{\shortstack{Single\\Data}}} & TimeGAN-NIPS2019  & 0.1607\textpm.000  & 0.1309\textpm.000 & 0.2037\textpm.006 & 0.1900\textpm.004 & 0.1768\textpm.002 & 0.1684\textpm.000 & \cellcolor{blue!10} \textbf{\underline{0.1783}}\textpm.005 & 0.1731\textpm.003 \\
\cline{3-11}
& & TimeVAE-arXiv2022  & 0.1606\textpm.000 & 0.1309\textpm.002 & 0.2039\textpm.005 & \cellcolor{blue!10} \textbf{\underline{0.1857}}\textpm.006 & 0.1782\textpm.003 & 0.1654\textpm.000 & 0.1785\textpm.001 & 0.1733\textpm.001\\
\cline{3-11}
& & DiffWave-ICLR2021 & 0.1606\textpm.002 & 0.1310\textpm.001 & 0.2039\textpm.003 & 0.1860\textpm.000 & \cellcolor{blue!10} \textbf{\underline{0.1768}}\textpm.002 & 0.1655\textpm.005 & 0.1784\textpm.003 & 0.1730\textpm.001 \\
\cline{3-11}
& & DiffTime-NIPS2023 & 0.1607\textpm.000 & 0.1310\textpm.000 & 0.2038\textpm.002 & 0.1861\textpm.001 & 0.1784\textpm.001 & 0.1654\textpm.000 & 0.1785\textpm.001 & 0.1733\textpm.000 \\
\cline{3-11}
& & Diffusion-TS-ICLR2024  & 0.1607\textpm.000 & 0.1309\textpm.000 & 0.2037\textpm.004 & 0.1864\textpm.002 & 0.1778\textpm.001 &  0.1654\textpm.001 & 0.1786\textpm.003 & 0.1787\textpm.002 \\
\cline{2-11}
   & \multirow{4}{*}{\textbf{\shortstack{Multi\\Data}}} & TimeDP-AAAI2025  & 0.1607\textpm.000 & 0.1310\textpm.000 & 0.2038\textpm.000 & 0.1864\textpm.000 & 0.1777\textpm.000 & 0.1655\textpm.000 & 0.1786\textpm.000 & 0.1735\textpm.000 \\
   \cline{3-11}
   &  & UPLOTS (Empty)  & 0.1614\textpm.001 & 0.1316\textpm.000 & 0.2132\textpm.002 & 0.1960\textpm.002 & 0.1865\textpm.001 & 0.1683\textpm.000 & 0.1873\textpm.001 & 0.1738\textpm.000 \\
   \cline{3-11}
   & & UPLOTS (GPT-2) & 0.1606\textpm.001 & \cellcolor{blue!10} \textbf{\underline{0.1309}}\textpm.000 & \cellcolor{blue!10} \textbf{\underline{0.2037}}\textpm.000 & 0.1865\textpm.001 & 0.1775\textpm.000 & \cellcolor{blue!10} \textbf{\underline{0.1653}}\textpm.000 & 0.1786\textpm.000 & \cellcolor{blue!10} \textbf{\underline{0.1729}}\textpm.000 \\
   \cline{3-11}
   & & UPLOTS (LLaMA) & \cellcolor{blue!10} \textbf{\underline{0.1605}}\textpm.000 & 0.1310\textpm.000 & \cellcolor{blue!10} \textbf{\underline{0.2037}}\textpm.001 & 0.1864\textpm.000  & 0.1775\textpm.000 & \cellcolor{blue!10} \textbf{\underline{0.1653}}\textpm.000 & 0.1786\textpm.000 & \cellcolor{blue!10} \textbf{\underline{0.1729}}\textpm.001 \\
   \cline{3-11}

\hline
\bottomrule
\end{tabular}
\end{table*}

%% file: table/main48.tex
\begin{table}[H]
\centering
\caption{Peak-pool comparison at generation length \(L=48\) on ETTh and Energy. Lower is better; bold and underlined values are best within each scenario.}
\label{tab:main48}
\tiny
\renewcommand{\arraystretch}{1.01}
\setlength{\tabcolsep}{0.75mm}
\begin{tabular}{@{}cclcccc@{}}
\hline
\textbf{Metric} & \textbf{Type} & \textbf{Model} & \textbf{ETTh\_MP} & \textbf{ETTh\_EP} & \textbf{Energy\_MP} & \textbf{Energy\_EP} \\
\hline
\multirow{9}{*}{\textbf{Context-FID}} & \multirow{5}{*}{\textbf{Single}} & TimeGAN & 0.9031\textpm.093 & 0.3231\textpm.066 & 0.8046\textpm.122 & 0.5361\textpm.037 \\
\cline{3-7}
 &  & TimeVAE & 0.8859\textpm.072 & 0.5697\textpm.067 & 0.5770\textpm.061 & 0.6614\textpm.073 \\
\cline{3-7}
 &  & DiffWave & 0.2156\textpm.047 & 0.1466\textpm.012 & 0.0969\textpm.009 & 0.0975\textpm.013 \\
\cline{3-7}
 &  & DiffTime & 0.2058\textpm.045 & 0.1214\textpm.009 & 0.0839\textpm.013 & 0.0872\textpm.009 \\
\cline{3-7}
 &  & Diffusion-TS & 0.2042\textpm.044 & 0.2123\textpm.035 & 0.0825\textpm.015 & 0.0760\textpm.007 \\
\cline{2-7}
 & \multirow{4}{*}{\textbf{Multi}} & TimeDP & 0.7200\textpm.197 & 0.1003\textpm.060 & 0.0369\textpm.005 & 0.0383\textpm.006 \\
\cline{3-7}
 &  & UPLOTS (Empty) & 0.2083\textpm.063 & 0.2599\textpm.046 & 0.3331\textpm.041 & 0.2445\textpm.024 \\
\cline{3-7}
 &  & UPLOTS (GPT-2) & 0.0878\textpm.009 & \cellcolor{blue!10} \textbf{\underline{0.0136}}\textpm.001 & 0.0348\textpm.004 & 0.0334\textpm.002 \\
\cline{3-7}
 &  & UPLOTS (LLaMA) & \cellcolor{blue!10} \textbf{\underline{0.0854}}\textpm.003 & 0.0145\textpm.002 & \cellcolor{blue!10} \textbf{\underline{0.0323}}\textpm.002 & \cellcolor{blue!10} \textbf{\underline{0.0315}}\textpm.002 \\
\hline
\multirow{9}{*}{\shortstack{\textbf{Discriminative}\\\textbf{Score}}} & \multirow{5}{*}{\textbf{Single}} & TimeGAN & 0.0492\textpm.059 & 0.0215\textpm.022 & 0.0534\textpm.039 & 0.0126\textpm.006 \\
\cline{3-7}
 &  & TimeVAE & 0.0462\textpm.039 & 0.0192\textpm.019 & 0.0361\textpm.030 & 0.0440\textpm.029 \\
\cline{3-7}
 &  & DiffWave & 0.0426\textpm.016 & 0.0194\textpm.021 & 0.0199\textpm.003 & 0.0293\textpm.018 \\
\cline{3-7}
 &  & DiffTime & 0.0395\textpm.015 & 0.0179\textpm.016 & 0.0147\textpm.004 & 0.0273\textpm.019 \\
\cline{3-7}
 &  & Diffusion-TS & 0.0941\textpm.115 & 0.0890\textpm.030 & 0.0373\textpm.017 & 0.0430\textpm.017 \\
\cline{2-7}
 & \multirow{4}{*}{\textbf{Multi}} & TimeDP & 0.0389\textpm.034 & 0.0324\textpm.028 & 0.0124\textpm.008 & 0.0297\textpm.025 \\
\cline{3-7}
 &  & UPLOTS (Empty) & 0.1105\textpm.074 & 0.1012\textpm.014 & 0.0921\textpm.008 & 0.0892\textpm.002 \\
\cline{3-7}
 &  & UPLOTS (GPT-2) & 0.0236\textpm.003 & \cellcolor{blue!10} \textbf{\underline{0.0136}}\textpm.005 & 0.0127\textpm.005 & 0.0238\textpm.004 \\
\cline{3-7}
 &  & UPLOTS (LLaMA) & \cellcolor{blue!10} \textbf{\underline{0.0227}}\textpm.005 & 0.0143\textpm.005 & \cellcolor{blue!10} \textbf{\underline{0.0116}}\textpm.003 & \cellcolor{blue!10} \textbf{\underline{0.0229}}\textpm.003 \\
\midrule
\multirow{9}{*}{\shortstack{\textbf{Predictive}\\\textbf{Score}}} & \multirow{5}{*}{\textbf{Single}} & TimeGAN & 0.1727\textpm.002 & 0.1308\textpm.000 & 0.2162\textpm.001 & 0.1858\textpm.000 \\
\cline{3-7}
 &  & TimeVAE & 0.1615\textpm.000 & 0.1307\textpm.000 & 0.2037\textpm.000 & \cellcolor{blue!10} \textbf{\underline{0.1853}}\textpm.000 \\
\cline{3-7}
 &  & DiffWave & 0.1617\textpm.000 & 0.1310\textpm.001 & 0.2036\textpm.000 & 0.1865\textpm.001 \\
\cline{3-7}
 &  & DiffTime & 0.1617\textpm.000 & 0.1309\textpm.000 & 0.2034\textpm.000 & 0.1863\textpm.000 \\
\cline{3-7}
 &  & Diffusion-TS & 0.1609\textpm.000 & 0.1309\textpm.000 & 0.2034\textpm.000 & 0.1861\textpm.000 \\
\cline{2-7}
 & \multirow{4}{*}{\textbf{Multi}} & TimeDP & 0.1616\textpm.000 & 0.1309\textpm.000 & 0.2035\textpm.000 & 0.1865\textpm.000 \\
\cline{3-7}
 &  & UPLOTS (Empty) & 0.1613\textpm.000 & 0.1312\textpm.000 & 0.2135\textpm.001 & 0.1967\textpm.001 \\
\cline{3-7}
 &  & UPLOTS (GPT-2) & 0.1615\textpm.001 & \cellcolor{blue!10} \textbf{\underline{0.1307}}\textpm.000 & \cellcolor{blue!10} \textbf{\underline{0.2033}}\textpm.000 & 0.1865\textpm.001 \\
\cline{3-7}
 &  & UPLOTS (LLaMA) & \cellcolor{blue!10} \textbf{\underline{0.1614}}\textpm.000 & 0.1308\textpm.000 & \cellcolor{blue!10} \textbf{\underline{0.2033}}\textpm.001 & 0.1864\textpm.000 \\
\hline
\end{tabular}
\end{table}

\begin{table}[H]
\centering
\caption{Peak-pool comparison at generation length \(L=48\) on PEMS04 and PEMS08. Lower is better; bold and underlined values are best within each scenario.}
\label{tab:main48_pems}
\tiny
\renewcommand{\arraystretch}{1.01}
\setlength{\tabcolsep}{0.75mm}
\begin{tabular}{@{}cclcccc@{}}
\hline
\textbf{Metric} & \textbf{Type} & \textbf{Model} & \textbf{PEMS04\_MP} & \textbf{PEMS04\_EP} & \textbf{PEMS08\_MP} & \textbf{PEMS08\_EP} \\
\hline
\multirow{9}{*}{\textbf{Context-FID}} & \multirow{5}{*}{\textbf{Single}} & TimeGAN & 0.0990\textpm.011 & 0.1058\textpm.017 & 0.3113\textpm.038 & 0.4196\textpm.039 \\
\cline{3-7}
 &  & TimeVAE & 0.7822\textpm.124 & 0.7798\textpm.086 & 0.8289\textpm.126 & 0.6162\textpm.059 \\
\cline{3-7}
 &  & DiffWave & 0.0941\textpm.006 & 0.0996\textpm.007 & 0.0845\textpm.012 & 0.1005\textpm.016 \\
\cline{3-7}
 &  & DiffTime & 0.0814\textpm.004 & 0.0895\textpm.010 & 0.0741\textpm.005 & 0.0942\textpm.012 \\
\cline{3-7}
 &  & Diffusion-TS & 0.0491\textpm.002 & 0.0221\textpm.004 & 0.0389\textpm.003 & 0.0212\textpm.001 \\
\cline{2-7}
 & \multirow{4}{*}{\textbf{Multi}} & TimeDP & 0.3669\textpm.099 & 0.3246\textpm.075 & 0.2694\textpm.047 & 0.1547\textpm.068 \\
\cline{3-7}
 &  & UPLOTS (Empty) & 0.8989\textpm.058 & 1.5606\textpm.155 & 0.7115\textpm.142 & 1.2696\textpm.085 \\
\cline{3-7}
 &  & UPLOTS (GPT-2) & 0.0188\textpm.003 & \cellcolor{blue!10} \textbf{\underline{0.0121}}\textpm.002 & \cellcolor{blue!10} \textbf{\underline{0.0152}}\textpm.001 & \cellcolor{blue!10} \textbf{\underline{0.0122}}\textpm.003 \\
\cline{3-7}
 &  & UPLOTS (LLaMA) & \cellcolor{blue!10} \textbf{\underline{0.0167}}\textpm.003 & 0.0134\textpm.005 & 0.0177\textpm.003 & 0.0143\textpm.002 \\
\hline
\multirow{9}{*}{\shortstack{\textbf{Discriminative}\\\textbf{Score}}} & \multirow{5}{*}{\textbf{Single}} & TimeGAN & 0.0052\textpm.003 & 0.0144\textpm.017 & 0.0319\textpm.025 & 0.0325\textpm.023 \\
\cline{3-7}
 &  & TimeVAE & 0.0393\textpm.025 & 0.0207\textpm.026 & 0.0289\textpm.024 & 0.0495\textpm.051 \\
\cline{3-7}
 &  & DiffWave & 0.0224\textpm.015 & 0.0184\textpm.024 & 0.0214\textpm.016 & 0.0201\textpm.021 \\
\cline{3-7}
 &  & DiffTime & 0.0182\textpm.013 & 0.0139\textpm.022 & 0.0191\textpm.013 & 0.0169\textpm.018 \\
\cline{3-7}
 &  & Diffusion-TS & 0.0374\textpm.003 & 0.0221\textpm.010 & 0.0229\textpm.005 & 0.0233\textpm.030 \\
\cline{2-7}
 & \multirow{4}{*}{\textbf{Multi}} & TimeDP & 0.0193\textpm.012 & 0.0296\textpm.025 & 0.0177\textpm.012 & 0.0201\textpm.008 \\
\cline{3-7}
 &  & UPLOTS (Empty) & 0.1563\textpm.068 & 0.2103\textpm.091 & 0.1749\textpm.017 & 0.2040\textpm.036 \\
\cline{3-7}
 &  & UPLOTS (GPT-2) & 0.0203\textpm.016 & \cellcolor{blue!10} \textbf{\underline{0.0129}}\textpm.007 & 0.0113\textpm.003 & 0.0125\textpm.003 \\
\cline{3-7}
 &  & UPLOTS (LLaMA) & \cellcolor{blue!10} \textbf{\underline{0.0184}}\textpm.011 & 0.0133\textpm.005 & \cellcolor{blue!10} \textbf{\underline{0.0111}}\textpm.002 & \cellcolor{blue!10} \textbf{\underline{0.0116}}\textpm.001 \\
\hline
\multirow{9}{*}{\shortstack{\textbf{Predictive}\\\textbf{Score}}} & \multirow{5}{*}{\textbf{Single}} & TimeGAN & \cellcolor{blue!10} \textbf{\underline{0.1759}}\textpm.000 & 0.1659\textpm.000 & 0.1800\textpm.000 & 0.1734\textpm.000 \\
\cline{3-7}
 &  & TimeVAE & 0.1776\textpm.000 & 0.1652\textpm.000 & 0.1783\textpm.000 & 0.1735\textpm.000 \\
\cline{3-7}
 &  & DiffWave & 0.1757\textpm.000 & 0.1652\textpm.000 & 0.1788\textpm.000 & 0.1732\textpm.002 \\
\cline{3-7}
 &  & DiffTime & 0.1757\textpm.000 & 0.1651\textpm.000 & 0.1785\textpm.000 & 0.1731\textpm.000 \\
\cline{3-7}
 &  & Diffusion-TS & 0.1773\textpm.001 & 0.1652\textpm.000 & 0.1788\textpm.000 & 0.1730\textpm.000 \\
\cline{2-7}
 & \multirow{4}{*}{\textbf{Multi}} & TimeDP & 0.1775\textpm.000 & 0.1653\textpm.000 & 0.1784\textpm.000 & 0.1731\textpm.000 \\
\cline{3-7}
 &  & UPLOTS (Empty) & 0.1855\textpm.001 & 0.1679\textpm.001 & 0.1874\textpm.000 & 0.1740\textpm.000 \\
\cline{3-7}
 &  & UPLOTS (GPT-2) & 0.1775\textpm.000 & 0.1651\textpm.000 & \cellcolor{blue!10} \textbf{\underline{0.1782}}\textpm.000 & \cellcolor{blue!10} \textbf{\underline{0.1727}}\textpm.000 \\
\cline{3-7}
 &  & UPLOTS (LLaMA) & 0.1775\textpm.000 & \cellcolor{blue!10} \textbf{\underline{0.1650}}\textpm.000 & \cellcolor{blue!10} \textbf{\underline{0.1782}}\textpm.000 & \cellcolor{blue!10} \textbf{\underline{0.1727}}\textpm.001 \\
\hline
\end{tabular}
\end{table}

%% file: table/main96.tex
\begin{table}[H]
\centering
\caption{Peak-pool comparison at generation length \(L=96\) on ETTh and Energy. Lower is better; bold and underlined values are best within each scenario.}
\label{tab:main96}
\tiny
\renewcommand{\arraystretch}{1.03}
\setlength{\tabcolsep}{0.75mm}
\begin{tabular}{@{}cclcccc@{}}
\hline
\textbf{Metric} & \textbf{Type} & \textbf{Model} & \textbf{ETTh\_MP} & \textbf{ETTh\_EP} & \textbf{Energy\_MP} & \textbf{Energy\_EP} \\
\hline
\multirow{9}{*}{\textbf{Context-FID}} & \multirow{5}{*}{\textbf{Single}} & TimeGAN & 1.3651\textpm.174 & 2.5919\textpm.332 & 1.4668\textpm.156 & 2.3660\textpm.428 \\
\cline{3-7}
 &  & TimeVAE & 0.9402\textpm.228 & 0.7119\textpm.119 & 0.7662\textpm.072 & 0.8743\textpm.067 \\
\cline{3-7}
 &  & DiffWave & 0.2954\textpm.075 & 0.1642\textpm.028 & 0.1512\textpm.019 & 0.1764\textpm.035 \\
\cline{3-7}
 &  & DiffTime & 0.2637\textpm.067 & 0.1560\textpm.021 & 0.1475\textpm.017 & 0.1521\textpm.022 \\
\cline{3-7}
 &  & Diffusion-TS & 0.2493\textpm.066 & 0.3479\textpm.084 & 0.1452\textpm.016 & 0.1312\textpm.007 \\
\cline{2-7}
 & \multirow{4}{*}{\textbf{Multi}} & TimeDP & 0.4402\textpm.082 & 0.2665\textpm.062 & 0.1962\textpm.056 & 0.2039\textpm.039 \\
\cline{3-7}
 &  & UPLOTS (Empty) & 0.2283\textpm.049 & 0.3537\textpm.073 & 0.4295\textpm.050 & 0.3270\textpm.017 \\
\cline{3-7}
 &  & UPLOTS (GPT-2) & \cellcolor{blue!10} \textbf{\underline{0.1335}}\textpm.013 & \cellcolor{blue!10} \textbf{\underline{0.0896}}\textpm.018 & 0.1443\textpm.025 & 0.1316\textpm.020 \\
\cline{3-7}
 &  & UPLOTS (LLaMA) & 0.1346\textpm.012 & 0.0900\textpm.008 & \cellcolor{blue!10} \textbf{\underline{0.1432}}\textpm.022 & \cellcolor{blue!10} \textbf{\underline{0.1311}}\textpm.015 \\
\hline
\multirow{9}{*}{\shortstack{\textbf{Discriminative}\\\textbf{Score}}} & \multirow{5}{*}{\textbf{Single}} & TimeGAN & 0.0839\textpm.061 & 0.2331\textpm.043 & 0.0437\textpm.043 & 0.0926\textpm.034 \\
\cline{3-7}
 &  & TimeVAE & 0.0248\textpm.033 & 0.0154\textpm.016 & 0.0328\textpm.033 & 0.0346\textpm.034 \\
\cline{3-7}
 &  & DiffWave & 0.0264\textpm.027 & 0.0168\textpm.015 & 0.0154\textpm.021 & 0.0201\textpm.016 \\
\cline{3-7}
 &  & DiffTime & 0.0235\textpm.024 & 0.0136\textpm.013 & 0.0126\textpm.022 & 0.0160\textpm.012 \\
\cline{3-7}
 &  & Diffusion-TS & 0.2075\textpm.049 & 0.0698\textpm.018 & 0.0438\textpm.017 & 0.0375\textpm.035 \\
\cline{2-7}
 & \multirow{4}{*}{\textbf{Multi}} & TimeDP & 0.0689\textpm.008 & 0.0512\textpm.012 & 0.0599\textpm.006 & 0.0608\textpm.002 \\
\cline{3-7}
 &  & UPLOTS (Empty) & 0.0947\textpm.078 & 0.1030\textpm.035 & 0.1021\textpm.005 & 0.1016\textpm.003 \\
\cline{3-7}
 &  & UPLOTS (GPT-2) & 0.0224\textpm.002 & \cellcolor{blue!10} \textbf{\underline{0.0111}}\textpm.003 & 0.0106\textpm.005 & 0.0136\textpm.003 \\
\cline{3-7}
 &  & UPLOTS (LLaMA) & \cellcolor{blue!10} \textbf{\underline{0.0197}}\textpm.005 & 0.0114\textpm.006 & \cellcolor{blue!10} \textbf{\underline{0.0105}}\textpm.003 & \cellcolor{blue!10} \textbf{\underline{0.0116}}\textpm.004 \\
\hline
\multirow{9}{*}{\shortstack{\textbf{Predictive}\\\textbf{Score}}} & \multirow{5}{*}{\textbf{Single}} & TimeGAN & 0.1744\textpm.001 & 0.1348\textpm.000 & 0.2131\textpm.002 & 0.2142\textpm.002 \\
\cline{3-7}
 &  & TimeVAE & 0.1603\textpm.000 & 0.1303\textpm.000 & 0.2033\textpm.000 & \cellcolor{blue!10} \textbf{\underline{0.1847}}\textpm.000 \\
\cline{3-7}
 &  & DiffWave & 0.1608\textpm.000 & 0.1304\textpm.000 & 0.2031\textpm.000 & 0.1858\textpm.001 \\
\cline{3-7}
 &  & DiffTime & 0.1607\textpm.000 & 0.1304\textpm.000 & 0.2030\textpm.000 & 0.1858\textpm.000 \\
\cline{3-7}
 &  & Diffusion-TS & 0.1607\textpm.000 & 0.1317\textpm.000 & 0.2033\textpm.000 & 0.1857\textpm.000 \\
\cline{2-7}
 & \multirow{4}{*}{\textbf{Multi}} & TimeDP & 0.1606\textpm.000 & 0.1309\textpm.000 & 0.2032\textpm.001 & 0.1858\textpm.001 \\
\cline{3-7}
 &  & UPLOTS (Empty) & 0.1611\textpm.000 & 0.1310\textpm.000 & 0.2138\textpm.000 & 0.1958\textpm.001 \\
\cline{3-7}
 &  & UPLOTS (GPT-2) & \cellcolor{blue!10} \textbf{\underline{0.1603}}\textpm.001 & \cellcolor{blue!10} \textbf{\underline{0.1303}}\textpm.000 & \cellcolor{blue!10} \textbf{\underline{0.2030}}\textpm.000 & 0.1859\textpm.001 \\
\cline{3-7}
 &  & UPLOTS (LLaMA) & \cellcolor{blue!10} \textbf{\underline{0.1603}}\textpm.000 & 0.1304\textpm.000 & 0.2031\textpm.001 & 0.1859\textpm.000 \\
\hline
\end{tabular}
\end{table}

\begin{table}[H]
\centering
\caption{Peak-pool comparison at generation length \(L=96\) on PEMS04 and PEMS08. Lower is better; bold and underlined values are best within each scenario.}
\label{tab:main96_pems}
\tiny
\renewcommand{\arraystretch}{1.03}
\setlength{\tabcolsep}{0.75mm}
\begin{tabular}{@{}cclcccc@{}}
\hline
\textbf{Metric} & \textbf{Type} & \textbf{Model} & \textbf{PEMS04\_MP} & \textbf{PEMS04\_EP} & \textbf{PEMS08\_MP} & \textbf{PEMS08\_EP} \\
\hline
\multirow{9}{*}{\textbf{Context-FID}} & \multirow{5}{*}{\textbf{Single}} & TimeGAN & 1.2871\textpm.083 & 0.3040\textpm.035 & 1.9413\textpm.231 & 2.5267\textpm.212 \\
\cline{3-7}
 &  & TimeVAE & 0.9719\textpm.157 & 0.7564\textpm.072 & 0.8939\textpm.093 & 0.4741\textpm.031 \\
\cline{3-7}
 &  & DiffWave & 0.1241\textpm.012 & 0.1314\textpm.011 & 0.1025\textpm.015 & 0.1120\textpm.008 \\
\cline{3-7}
 &  & DiffTime & 0.1020\textpm.009 & 0.1208\textpm.009 & 0.0942\textpm.011 & 0.0927\textpm.010 \\
\cline{3-7}
 &  & Diffusion-TS & 0.1016\textpm.015 & \cellcolor{blue!10} \textbf{\underline{0.0376}}\textpm.005 & 0.1010\textpm.012 & 0.0506\textpm.004 \\
\cline{2-7}
 & \multirow{4}{*}{\textbf{Multi}} & TimeDP & 0.7131\textpm.114 & 0.5951\textpm.096 & 0.5049\textpm.069 & 0.3950\textpm.120 \\
\cline{3-7}
 &  & UPLOTS (Empty) & 1.8241\textpm.118 & 2.7541\textpm.394 & 1.2804\textpm.150 & 2.2246\textpm.160 \\
\cline{3-7}
 &  & UPLOTS (GPT-2) & 0.0423\textpm.011 & 0.0725\textpm.013 & \cellcolor{blue!10} \textbf{\underline{0.0436}}\textpm.014 & \cellcolor{blue!10} \textbf{\underline{0.0406}}\textpm.011 \\
\cline{3-7}
 &  & UPLOTS (LLaMA) & \cellcolor{blue!10} \textbf{\underline{0.0412}}\textpm.009 & 0.0712\textpm.021 & 0.0446\textpm.016 & 0.0413\textpm.015 \\
\hline
\multirow{9}{*}{\shortstack{\textbf{Discriminative}\\\textbf{Score}}} & \multirow{5}{*}{\textbf{Single}} & TimeGAN & 0.0393\textpm.030 & 0.0252\textpm.020 & 0.0457\textpm.095 & 0.1023\textpm.030 \\
\cline{3-7}
 &  & TimeVAE & 0.0577\textpm.035 & 0.0438\textpm.049 & 0.0346\textpm.009 & 0.0329\textpm.028 \\
\cline{3-7}
 &  & DiffWave & 0.0196\textpm.011 & 0.0145\textpm.005 & 0.0175\textpm.005 & 0.0184\textpm.020 \\
\cline{3-7}
 &  & DiffTime & 0.0183\textpm.007 & 0.0123\textpm.003 & 0.0139\textpm.004 & 0.0147\textpm.019 \\
\cline{3-7}
 &  & Diffusion-TS & 0.0421\textpm.007 & 0.0150\textpm.007 & 0.0327\textpm.022 & 0.0136\textpm.023 \\
\cline{2-7}
 & \multirow{4}{*}{\textbf{Multi}} & TimeDP & 0.0753\textpm.015 & 0.0239\textpm.005 & 0.0264\textpm.011 & 0.0195\textpm.008 \\
\cline{3-7}
 &  & UPLOTS (Empty) & 0.1865\textpm.085 & 0.2283\textpm.097 & 0.1646\textpm.071 & 0.1661\textpm.083 \\
\cline{3-7}
 &  & UPLOTS (GPT-2) & 0.0101\textpm.004 & 0.0091\textpm.004 & 0.0136\textpm.004 & 0.0121\textpm.001 \\
\cline{3-7}
 &  & UPLOTS (LLaMA) & \cellcolor{blue!10} \textbf{\underline{0.0092}}\textpm.002 & \cellcolor{blue!10} \textbf{\underline{0.0085}}\textpm.003 & \cellcolor{blue!10} \textbf{\underline{0.0131}}\textpm.001 & \cellcolor{blue!10} \textbf{\underline{0.0116}}\textpm.002 \\
\hline
\multirow{9}{*}{\shortstack{\textbf{Predictive}\\\textbf{Score}}} & \multirow{5}{*}{\textbf{Single}} & TimeGAN & 0.1788\textpm.001 & 0.1656\textpm.001 & 0.1802\textpm.001 & 0.1782\textpm.001 \\
\cline{3-7}
 &  & TimeVAE & 0.1781\textpm.000 & 0.1650\textpm.000 & 0.1786\textpm.000 & 0.1740\textpm.000 \\
\cline{3-7}
 &  & DiffWave & 0.1759\textpm.000 & 0.1647\textpm.000 & 0.1786\textpm.000 & 0.1732\textpm.002 \\
\cline{3-7}
 &  & DiffTime & \cellcolor{blue!10} \textbf{\underline{0.1757}}\textpm.000 & 0.1646\textpm.000 & 0.1786\textpm.000 & 0.1731\textpm.000 \\
\cline{3-7}
 &  & Diffusion-TS & 0.1776\textpm.000 & 0.1648\textpm.000 & 0.1797\textpm.000 & 0.1731\textpm.000 \\
\cline{2-7}
 & \multirow{4}{*}{\textbf{Multi}} & TimeDP & 0.1776\textpm.000 & 0.1647\textpm.000 & 0.1789\textpm.000 & 0.1736\textpm.001 \\
\cline{3-7}
 &  & UPLOTS (Empty) & 0.1854\textpm.001 & 0.1678\textpm.000 & 0.1878\textpm.001 & 0.1742\textpm.000 \\
\cline{3-7}
 &  & UPLOTS (GPT-2) & 0.1775\textpm.000 & \cellcolor{blue!10} \textbf{\underline{0.1645}}\textpm.000 & \cellcolor{blue!10} \textbf{\underline{0.1783}}\textpm.000 & \cellcolor{blue!10} \textbf{\underline{0.1728}}\textpm.000 \\
\cline{3-7}
 &  & UPLOTS (LLaMA) & 0.1775\textpm.000 & \cellcolor{blue!10} \textbf{\underline{0.1645}}\textpm.000 & \cellcolor{blue!10} \textbf{\underline{ 0.1783}}\textpm.000 & \cellcolor{blue!10} \textbf{\underline{0.1728}}\textpm.001 \\
\hline
\end{tabular}
\end{table}